%% file: main.tex
\documentclass[10pt,twocolumn,letterpaper]{article}

\usepackage{iccv}
\usepackage{times}
\usepackage{epsfig}
\usepackage{graphicx}
\usepackage{amsmath}
\usepackage{amssymb}
\usepackage{afterpage}

\usepackage{booktabs}
\usepackage{xcolor}
\usepackage{color}
\usepackage{bbding}
\usepackage{array,multirow,textcomp}
\usepackage{rotating}
\usepackage[skip=0.5\baselineskip]{caption}
\usepackage{makecell}
\usepackage{booktabs}
\usepackage{adjustbox}
\usepackage{array}
\usepackage{balance}
\usepackage{wrapfig}
\usepackage{enumitem}
\usepackage{tikz}
\usepackage{tabularx}
\usepackage{colortbl}
\usepackage{placeins}
\usepackage{float}

\usepackage[hashEnumerators,hybrid]{markdown}
\usepackage{pgf}
\usepackage{subcaption}

\usepackage[pagebackref=true,breaklinks=true,colorlinks,bookmarks=false]{hyperref}
\usepackage[capitalize]{cleveref}

\iccvfinalcopy 

\def\iccvPaperID{9261} 

\crefname{section}{Sec.}{Secs.}
\Crefname{section}{Section}{Sections}
\Crefname{table}{Table}{Tables}
\crefname{table}{Tab.}{Tabs.}

\makeatletter
\renewcommand{\paragraph}{%
  \@startsection{paragraph}{4}%
  {\z@}{1.0ex \@plus 1ex \@minus .2ex}{-1em}%
  {\normalfont\normalsize\bfseries}%
}
\makeatother

\newcommand{\B}[1]{\textbf{#1}}
\DeclareMathOperator*{\argmax}{arg\,max}

\newcommand{\sem}{\mathrm{sem}}

\def\quads{\hskip0.5em\relax}
\newcolumntype{Y}{>{\centering\arraybackslash}X}
\newcolumntype{C}[1]{>{\centering\let\newline\\\arraybackslash\hspace{0pt}}m{#1}}
\newcolumntype{R}[2]{%
    >{\adjustbox{angle=#1,lap=\width-(#2)}\bgroup}%
    l%
    <{\egroup}%
}
\newcommand*\rots{\multicolumn{1}{R{60}{1em}}}

\newcommand{\spm}[1]{\tiny{$\,\pm$#1}}

\begin{document}

\title{EDAPS: Enhanced Domain-Adaptive Panoptic Segmentation}

\author{
    Suman Saha\thanks{These authors contributed equally to this work.} \\
    ETH Zurich
    \and
    Lukas Hoyer$^*$ \\
    ETH Zurich
    \and
    Anton Obukhov \\
    ETH Zurich
    \and
    Dengxin Dai \\
    Huawei Technologies
    \and
    Luc Van Gool \\
    ETH Zurich, KU Leuven
}

\maketitle

\input{text/abstract}

\input{text/intro}

\input{text/rel_work}
\input{text/method}

\input{text/exp}
\input{text/conclusion}

\clearpage

{\small
\bibliographystyle{ieee_fullname}
\bibliography{egbib}
}

\clearpage

\noindent\textbf{\Large Supplementary Material}

\makeatletter
\renewcommand{\theHsection}{papersection.\number\value{section}} 
\renewcommand{\thesection}{\Alph{section}}
\renewcommand{\thefigure}{S\arabic{figure}}
\renewcommand{\thetable}{S\arabic{table}}
\setcounter{section}{0}

\setcounter{figure}{0}
\setcounter{table}{0}
\makeatother

\input{text/sup_mat}

\end{document}

%% file: text/abstract.tex
\begin{abstract}
With autonomous industries on the rise, domain adaptation of the visual perception stack is an important research direction due to the cost savings promise.
Much prior art was dedicated to domain-adaptive semantic segmentation in the synthetic-to-real context.
Despite being a crucial output of the perception stack, panoptic segmentation has been largely overlooked by the domain adaptation community.
Therefore, we revisit well-performing domain adaptation strategies from other fields, adapt them to panoptic segmentation, and show that they can effectively enhance panoptic domain adaptation.
Further, we study the panoptic network design and propose a novel architecture (EDAPS) designed explicitly for domain-adaptive panoptic segmentation. It uses a shared, domain-robust transformer encoder to facilitate the joint adaptation of semantic and instance features, but task-specific decoders tailored for the specific requirements of both domain-adaptive semantic and instance segmentation. As a result, the performance gap seen in challenging panoptic benchmarks is substantially narrowed. EDAPS significantly improves the state-of-the-art performance for panoptic segmentation UDA by a large margin of 20\% on SYNTHIA-to-Cityscapes and even 72\% on the more challenging SYNTHIA-to-Mapillary Vistas.
The implementation is available at {\footnotesize\url{https://github.com/susaha/edaps}}.
\end{abstract}

%% file: text/intro.tex
\vspace{-10mm}
\section{Introduction}
\label{sec:intro}
Panoptic segmentation~\cite{kirillov2019panoptic} of images is a core computer vision task that jointly solves two related problems -- semantic segmentation and instance segmentation. 
With the rise of robotics and emerging autonomous driving markets, efficient visual perception stacks are in high demand.
However, large-scale supervised learning of panoptic segmentation%
~\cite{cheng2019panoptic,kirillov2019panoptic,kirillov2019panopticFP,porzi2019seamless,xiong2019upsnet,mohan2021efficientps} 
is prohibitively expensive as it requires dense annotations for both semantics and instances, which require time-consuming manual labeling.

A promising alternative to circumvent this issue is  
to learn from abundantly available photo-realistic synthetic images~\cite{ros2016synthia,richter2016playing} as their ground truth annotations can be automatically generated by the rendering engine.
However, often models trained on synthetic data (\emph{source domain})
fail to generalize well on the real data (\emph{target domain}) due to differences in data distribution, known as the \textit{domain gap}.

\begin{figure}[t]
  \centering
   \includegraphics[width=1.\linewidth]{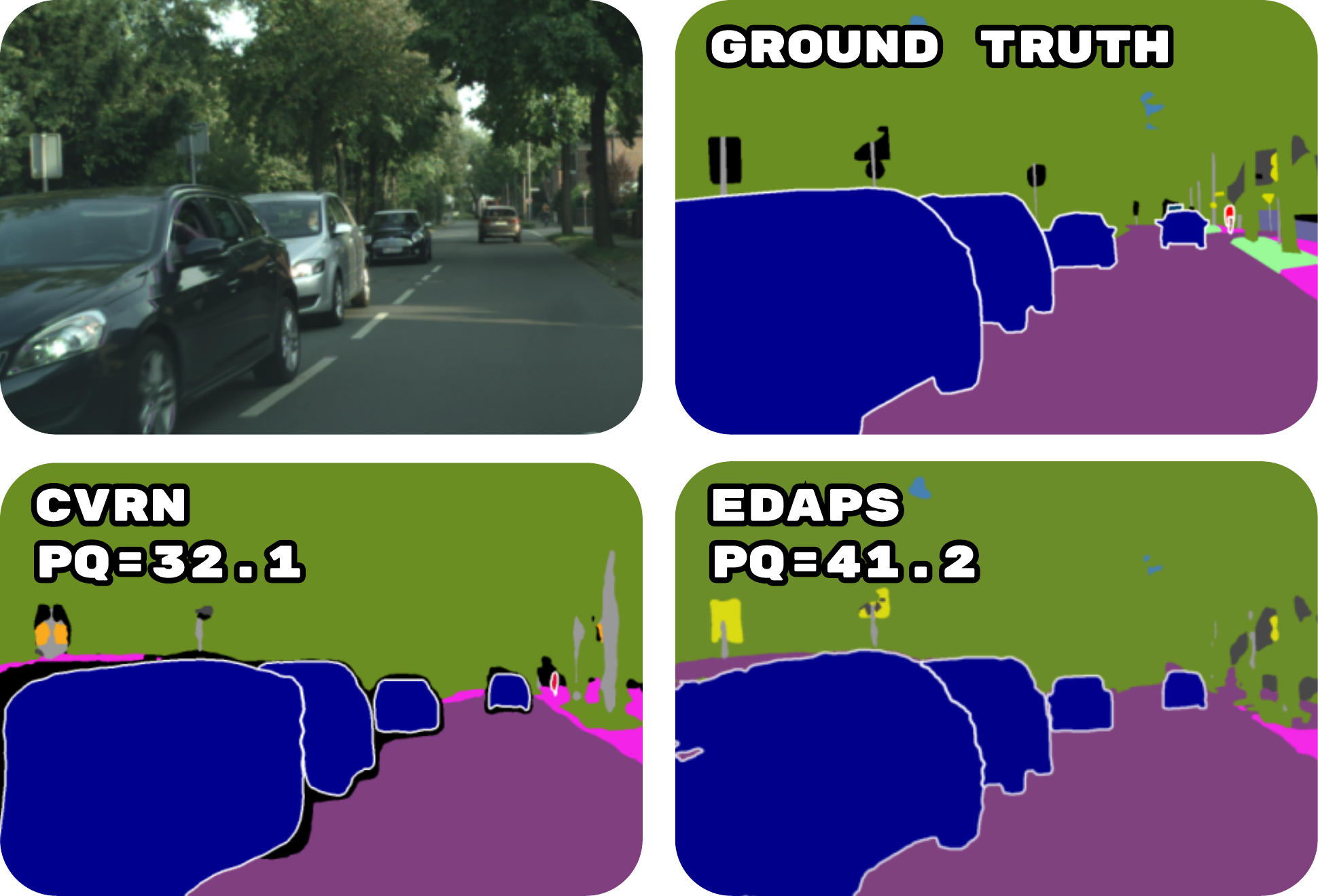}
   \caption{
   EDAPS is an architecture and a collection of recipes, designed specifically for Domain-Adaptive Panoptic Segmentation. 
   It demonstrates a significant improvement over the prior art on the challenging synthetic-to-real benchmarks.
   As shown above, it is better than CVRN~\cite{huang2021cross} on SYNTHIA $\rightarrow$ Cityscapes by $9$ mPQ. 
   }
   \label{fig:teaser}
\end{figure}

A common remedy to this problem is to minimize the domain gap using Unsupervised Domain Adaptation (UDA).
This field is actively studied for 
image classification~\cite{
long2015learning,ganin2016domain,long2018conditional,saito2018maximum,pan2019transferrable}, 
object detection~\cite{
chen2018domain,saito2019strong, xu2020cross, chen2021scale, li2022cross}, 
and semantic segmentation~\cite{
hoffman2016fcns,tsai2018learning,hoffman2018cycada,li2019bidirectional,saha2021learning,tranheden2021dacs,hoyer2021daformer,hoyer2022hrda,hoyer2023domain}.
However, UDA for panoptic segmentation is often overlooked and there are only two works, namely CVRN~\cite{huang2021cross} and UniDAPS~\cite{zhang2022hierarchical}, which address this problem from the synthetic-to-real point of view. 
Compared to the related semantic segmentation UDA, these approaches achieve only subpar performance. 
Specifically, the relative performance of the best UDA and fully-supervised learning approaches to panoptic segmentation (64\% in UniDAPS~\cite{zhang2022hierarchical}) is much smaller than that of semantic segmentation (88\% in DAFormer~\cite{hoyer2021daformer}).

To understand the root cause of the identified performance gap, we revisit the progress of UDA in semantic segmentation, lift the well-performing UDA strategies to panoptic segmentation, and show that they can effectively enhance panoptic segmentation UDA.

\begin{figure}[t]
  \centering
  \includegraphics[width=1.0\linewidth]{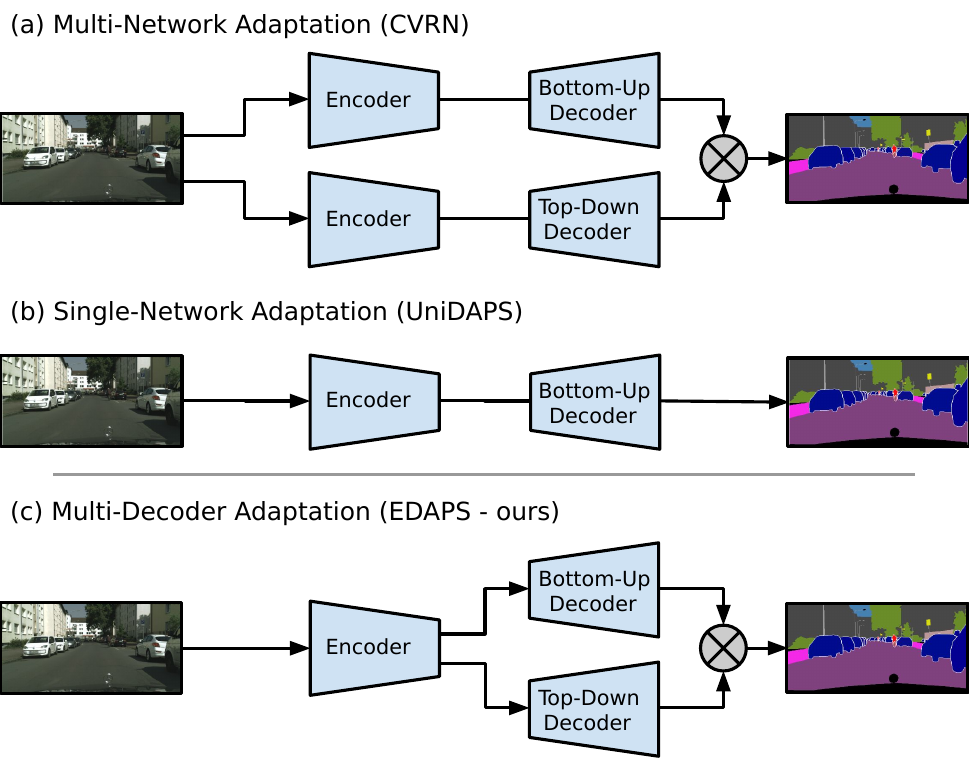}
  \caption{
    Overview of network architectures employed in prior panoptic UDA works and the proposed architecture. CVRN~\cite{huang2021cross} 
    (a) does not share parameters between panoptic branches, whereas UniDAPS~\cite{zhang2022hierarchical} 
    (b) resorts to the opposite extreme and shares everything. 
    With EDAPS (c), we propose to share the encoder but to use task-specific decoders, which facilitates panoptic domain adaptation.
  }
  \label{fig:architecture_comparison}
\end{figure}

Further, we revisit the panoptic network design and conduct a study of principal architecture designs for panoptic segmentation with respect to their UDA capabilities. We show that previous UDA methods took sub-optimal design choices. While separating the networks for both tasks prevents the network from jointly adapting task-shared features from the source to the target domain (see Fig.~\ref{fig:architecture_comparison}\,a), a shared encoder-decoder cannot accommodate the different needs (such as task-specific knowledge or architecture design) when adapting both tasks  (see Fig.~\ref{fig:architecture_comparison}\,b).

To address these problems, we propose EDAPS, a network architecture that is particularly designed for domain-adaptive panoptic segmentation. It uses a shared transformer~\cite{vaswani2017attention} encoder to facilitate the joint adaptation of semantic and instance features, but task-specific decoders tailored for the specific requirements of 
both domain-adaptive semantic segmentation and domain-adaptive instance segmentation (see Fig.~\ref{fig:architecture_comparison}\,c). Specifically, separate decoders can learn task-specific parameters, mitigating the issue of conflicting gradients~\cite{yu2020gradient}, and their architectures can be independently chosen to best suit semantic or instance segmentation. Even though this design was used in the supervised setting, we are the first to identify its strong potential specifically for panoptic UDA.

Utilizing the enhanced panoptic UDA and the enhanced domain-adaptive panoptic network design, EDAPS shows significant performance gains over the prior works on a number of standard perception benchmarks (Fig.~\ref{fig:teaser}).
EDAPS improves the state-of-the-art mPQ from $33.0$ to $41.2$ on On SYNTHIA $\rightarrow$ Cityscapes and from $21.3$ to $36.6$ on SYNTHIA $\rightarrow$ Mapillary Vistas. 
EDAPS trains semantic segmentation and instance segmentation tasks together using a joint training optimization. Therefore, EDAPS can be trained end-to-end in a single stage and the training only requires 21 hours on a single RTX 2080 Ti GPU, which improves its applicability for the community.

The main contribution of EDAPS is the novel combination of network components and UDA strategies on a system level, which results in a $major$ relative gain in the state-of-the-art panoptic segmentation UDA performance of 20\% on SYNTHIA-to-Cityscapes and even 72\% on the more challenging SYNTHIA-to-Mapillary Vistas.
In particular, we carefully study various principal panoptic architectures for their UDA capability and identify a network design that is particularly suited for panoptic UDA. It improves the UDA performance over strong baselines%
, while being parameter-efficient and fast at inference.
Further, this is the first paper that lifts recent UDA techniques to panoptic segmentation and systematically studies their effectiveness for panoptic UDA.

%% file: text/rel_work.tex
\section{Related Work}
\label{sec:relwork}
We outline two cornerstone classes of works related to UDA in Semantic and Panoptic Segmentation settings.

\paragraph{UDA for Semantic Segmentation}
Methods in this class take input images from source and target domains along with the source semantic ground truth label.
A supervised cross-entropy loss is computed on the source image semantic prediction. 
A UDA semantic loss (e.g., adversarial \cite{vu2019advent} or pseudo-label-based self-training \cite{zou2018unsupervised} loss) is used for domain alignment that operates on the semantic feature space. 
UDA for semantic segmentation is one of the most popular tasks in dense prediction, and a large number of works in this category exist in the literature~\cite{
vu2019advent,tsai2018learning,hoffman2018cycada,li2019bidirectional,kim2020learning,zhou2021context,tranheden2021dacs,melas2021pixmatch,choi2019self,araslanov2021self,zhang2021prototypical,zou2018unsupervised,yang2020fda,wang2020differential,luo2019taking,du2019ssf,chen2018road,hoffman2016fcns,kang2020pixel,zhang2019category,liu2021learning,saha2021learning}. 
The first approach of using an adversarial loss helps to align the source and target domain distributions at the input~\cite{
hoffman2018cycada,gong2021dlow}, 
at the feature-level~\cite{
tsai2018learning,hoffman2016fcns}, 
at the output~\cite{
tsai2018learning,vu2019advent},
or at the patch level~\cite{tsai2019domain}.
In the second approach of using self-training~\cite{
zou2018unsupervised,li2019bidirectional,zhang2019category,zou2019confidence}, 
pseudo-labels are generated on the unannotated target domain images either offline~\cite{
sakaridis2018model,yang2020fda,zou2018unsupervised,zou2019confidence}
or online~\cite{
zhang2021prototypical,wang2021domain,tranheden2021dacs,zhou2021context,hoyer2021daformer,hoyer2022hrda}.
The pseudo-labels can be stabilized with consistency regularization~\cite{
tarvainen2017mean,sohn2020fixmatch} based on pseudo-label prototypes~\cite{zhang2021prototypical}, different data augmentation schemes~\cite{
araslanov2021self,choi2019self,melas2021pixmatch,hoyer2023mic}, cross-domain mixup strategies~\cite{
tranheden2021dacs,zhou2021context}, and multiple resolutions~\cite{hoyer2022hrda,hoyer2023domain}.

\paragraph{UDA for Panoptic Segmentation}
Considering the used network architectures, there are distinct shortcomings in both CVRN~\cite{huang2021cross} and UniDAPS~\cite{zhang2022hierarchical}.
On the one side, CVRN requires the training of two separate networks for semantic segmentation and instance segmentation, which is time-consuming, parameter-inefficient, and slow during inference (Fig.~\ref{fig:architecture_comparison}). 
Furthermore, no task-agnostic knowledge can be shared during the adaptation process by separating the semantic from the instance network.
The approach relies on expensive multi-stage training (i.e., training, pseudo-label generation, and retraining with fused labels).
On the other side, UniDAPS proposed to adapt panoptic segmentation within a unified network~\cite{carion2020end}, which directly predicts panoptic segments instead of separate semantic and instance masks.
While this simple concept is intriguing and performs well in a supervised setting~\cite{carion2020end, cheng2021per, li2022panoptic}, unified network architectures are not inherently suited for UDA as shown in~\cite{zhang2022hierarchical}. Even with specific UDA strategies to compensate for that, UniDAPS achieves only a small improvement over CVRN.

The authors of \cite{hoyer2021daformer,hoyer2023domain} pointed out that architectures and training schemes that work for supervised learning might not work as desirably for UDA, and special care is required (in network design and training recipes) to address the domain shift problem effectively.
Outside of the synthetic-to-real domain, the authors of~\cite{pdam} proposed a panoptic UDA approach to microscopy image analysis.
In this paper, we propose a domain-adaptive panoptic segmentation framework that carefully selects the network design and training recipes tailored explicitly for UDA.

%% file: text/method.tex
\vspace{-2mm}
\section{Method}
\label{sec:method}

In this section, we first recap panoptic image segmentation. Second, we present our enhanced panoptic segmentation UDA pipeline. Third, we define principal panoptic architectures for a systematic analysis of their UDA capabilities. And finally, we introduce our EDAPS network architecture, specifically designed for panoptic UDA.

\subsection{Supervised Panoptic Segmentation}

Panoptic segmentation is commonly approached by decomposing the task into a semantic segmentation and an instance segmentation component so that the panoptic segmentation loss $\mathcal{L}_\mathrm{PS}$ is composed of a semantic and instance loss. $\mathcal{L}_\mathrm{PS} = \mathcal{L}_\sem + \mathcal{L}_\mathrm{inst}$. The semantic segmentation loss typically uses a pixel-wise cross-entropy loss to assign each pixel of an image to one class from a pre-defined set.
Instance segmentation further distinguishes instances within classes with countable entities, the so-called thing-classes such as \emph{car} or \emph{person} (as opposed to uncountable stuff-classes such as \emph{road} or \emph{sky}). Instance segmentation can be approached in two ways. In top-down approaches such as Mask R-CNN~\cite{he2017mask}, instances are predicted based on proposals in the form of bounding boxes and instances masks. In bottom-up approaches such as Panoptic-DeepLab~\cite{cheng2019panoptic}, the instances are grouped on a pixel-level without proposals, for example using pixel-wise instance center and offset heatmaps.
During inference, semantic and instance segmentations are deterministically fused into a panoptic segmentation representation.

\begin{figure}[t]
  \centering
  \includegraphics[width=\linewidth]{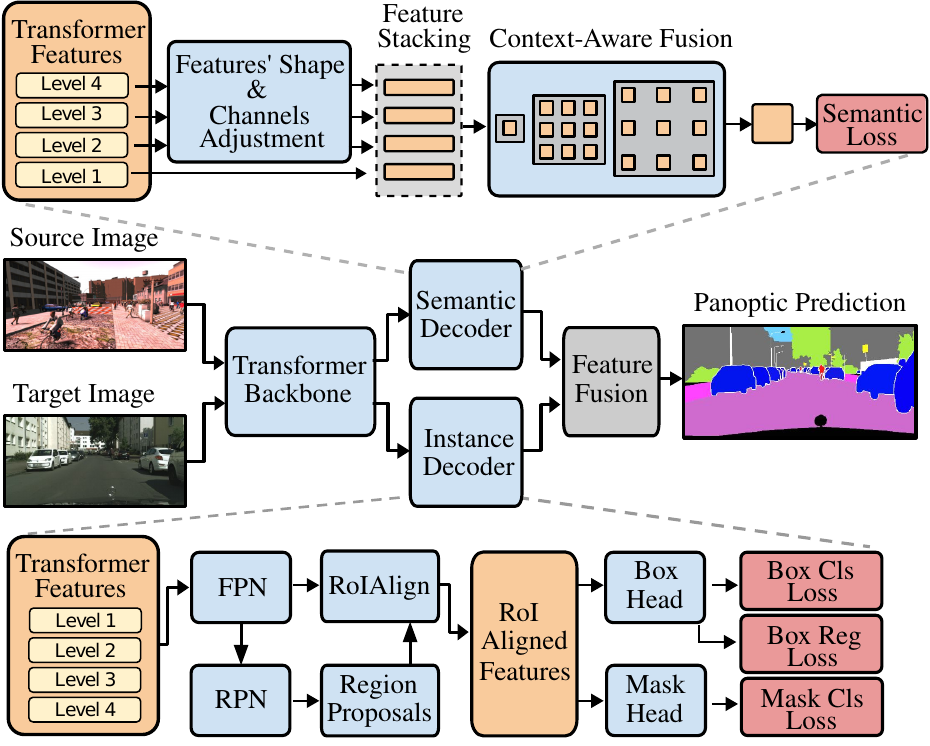}
  \caption{
    The proposed EDAPS (Enhanced Domain-Adaptive Panoptic Segmentation) network architecture.
    EDAPS is built with many design choices tailored to UDA in mind. 
    It achieves competitive results on challenging synthetic-to-real panoptic segmentation benchmarks.
  }
  \label{fig:overview}
\end{figure}

\vspace{-1mm}
\subsection{Enhanced Panoptic UDA}
In the UDA setting, a neural network $\mathcal{F}_{\theta}$ is trained on annotated source domain images 
$\mathcal{X}^{(s)} = \{x_i^{(s)}\}_{i=1}^{N^{(s)}}$ 
and unannotated target domain images 
$\mathcal{X}^{(t)} = \{x_i^{(t)}\}_{i=1}^{N^{(t)}}$ 
with $x_i^{(s)}, x_i^{(t)} \in \mathbb{R}^{H \times W \times 3}$
with the objective of achieving good performance on the target domain.
As panoptic segmentation ground truth $\mathcal{Y}^{(s)} = \{y^{(s)}_i\}_{i=1}^{N_S}$ is only available on the source domain, the panoptic model can only be trained with source data in a supervised fashion. In particular, the source loss term $\mathcal{L}^{(s)} = \mathcal{L}_\mathrm{PS}(\hat{y}^{(s)}, y^{(s)})$.

Models trained using the supervised loss $\mathcal{L}^{(s)}$ on the source domain often exhibit poor generalization on the target domain due to the ``domain gap'' between the distributions of source and target data.
To adapt the model to the target domain, an additional unsupervised loss term $\mathcal{L}^{(t)}$ is computed on the target domain.
The overall loss combines both the source and target loss terms $\mathcal{L}_\mathrm{UDA} = \mathcal{L}^{(s)} + \mathcal{L}^{(t)}$.

Various unsupervised target loss terms were proposed in the literature, mostly based on adversarial training~\cite{tsai2018learning,tsai2019domain,wang2020classes} or self-training~\cite{zou2018unsupervised,zhang2019category,mei2020instance,tranheden2021dacs,zhang2021prototypical,hoyer2021improving}.

\paragraph{Self-Training on Target Domain}
In this work, we resort to self-training for adapting the network to the target domain similar to previous panoptic UDA methods~\cite{xu2020cross, zhang2022hierarchical}. In self-training, the model $\mathcal{F}_{\theta}$ is trained with high-confidence pseudo-labels~\cite{lee2013pseudo} on the target domain using a weighted cross-entropy loss
\begin{equation}
  \mathcal{L}^{(t)} = -\sum_{i,j,c} 
  q^{(t)}_{i,j}
  \left( 
      p^{(t)} \log(\hat{y}^{(t)})
  \right)_{i,j,c}
\end{equation}
which helps the network gradually adapt to the target domain. Here, $p^{(t)}$ denotes the pseudo-label and $q^{(t)}$ its confidence estimate.

The pseudo-labels can be generated using the predictions of a teacher network $\mathcal{T}_{\phi}$. 
As the output of $\mathcal{T}_\phi$ is a map of per-pixel probabilities, a single class assignment is done using the mode of categorical distribution, which could be converted back to one-hot categorical form via the Iverson bracket $[\cdot]$:
\begin{equation}
    p_\mathit{i,j,c}^{(t)} = \left[
      c = \argmax_{c'} \mathcal{T}_\phi(x^{(t)})_{i,j,c'}
    \right]\,.
\end{equation}
Since the pseudo-labels are the predictions of a network,
they are not always correct, and their quality can be estimated by a confidence estimate $q^{(t)}$~\cite{zou2018unsupervised,mei2020instance,tranheden2021dacs,hoyer2021improving} 
\begin{equation}
    q^{(t)} = \frac{\sum_{i,j} \left[\max_{c'} \mathcal{T}_\phi(x^{(t)})_{(ijc')} > \tau\right]}{H \cdot W}\,.
\end{equation}

To stabilize the quality of the pseudo-labels, we resort to the mean teacher framework~\cite{tarvainen2017mean}, which is commonly used in semantic segmentation UDA~\cite{araslanov2021self,tranheden2021dacs,liu2021bapa,hoyer2021daformer}. The parameters of the teacher network $\mathcal{T}_{\phi}$ are updated with the exponential moving average of the parameters of the student network $\mathcal{F}_\theta$ at every training step $t$:
\begin{equation}
    \phi_{t+1} \leftarrow \alpha \phi_t + (1 - \alpha) \theta_t\,.
\end{equation}

Alongside self-training, we adopt a consistency training~\cite{sajjadi2016regularization, tarvainen2017mean, sohn2020fixmatch},
in which the student network $\mathcal{F}_\theta$ is trained on augmented target images,
whereas, the mean teacher network $\mathcal{F}_\phi$ predicts pseudo-labels for actual target images.
The augmented images are generated by mixing pixels of the source and target images.
The mixing is done following the Class-Mix strategy~\cite{olsson2021classmix}, which has been successfully applied to semantic segmentation UDA~\cite{tranheden2021dacs,liu2021bapa,hoyer2021daformer}: first, we randomly select $N/2$ semantic classes among $N$ classes present in the source image.
Next, the pixels belonging to these selected $N/2$ semantic classes are pasted into the target image, resulting in a new augmented image.

Furthermore, we leverage the recent findings in semantic segmentation UDA~\cite{hoyer2021improving} to further enhance the performance of the panoptic UDA.
More specifically, we adopt rare class sampling and a feature regularization loss based on ImageNet features to learn domain-invariant representations for panoptic UDA.

\paragraph{Rare Class Sampling (RCS)}
Most datasets are imbalanced, so certain classes are underrepresented. It can hurt the adaptation process if a certain class is sampled with a low frequency~\cite{hoyer2021daformer}. Therefore, we follow DAFormer~\cite{hoyer2021daformer} and sample images with rare classes more frequently. The sampling probability of class $c$ is
\begin{equation}
    P(c) = \frac{e^{(1-f_c) / T}}{\sum_{c'=1}^C e^{(1-f_{c'}) / T}}\,.
    \label{eq:P_c}
\end{equation}
where $f_c$ denotes the class frequency in the source dataset and $T$ is a temperature parameter. Given a sampled class $c \sim P$, a source sample is drawn from the subset containing class $c$: $x^{(s)} \sim \text{uniform}(\mathcal{X}^{(s,c)})$.

\paragraph{ImageNet Feature Distance (FD)}
The encoder network in panoptic segmentation UDA is usually pre-trained using ImageNet classification. 
Given that ImageNet is a real-world dataset, its features can be valuable when adapting to a real-world target domain. 
However, it was observed that some relevant features could be forgotten during the adaptation process~\cite{hoyer2021daformer}. 
Therefore, it can be useful to regularize the adaptation process with an ImageNet feature distance loss term~\cite{hoyer2021daformer}
\begin{equation}
    \mathcal{L}_\mathit{FD} = \left\lVert \mathcal{F}_\mathrm{ImageNet}(x^{(s)}) - \mathcal{F}_\theta(x^{(s)})\right\rVert_2\,,
\end{equation}
where $F_\mathrm{ImageNet}$ denotes the frozen ImageNet model. 
As ImageNet annotates thing classes, it is beneficial to constraint the feature distance to image regions that are labeled as thing classes~\cite{hoyer2021daformer}.

\subsection{Principal Panoptic Architectures for UDA}
\label{sec:methods_principal_architectures}

To further facilitate panoptic UDA, we aim to design a network architecture particularly tailored for domain-adaptive panoptic segmentation.
Several different network architectures have been proposed for panoptic segmentation in the supervised setting. However, the effect of the different designs on the performance in a domain adaptation setting is not well studied. Therefore, we analyze four principal panoptic architectures in a fair comparison using the same network building blocks and the same enhanced panoptic UDA strategy in \S\ref{sec:network_study}. The systematic comparison is the foundation for our EDAPS architecture.

\paragraph{M-Net}

uses two separate encoder-decoder networks for semantic and instance segmentation (see Fig.~\ref{fig:architecture_comparison}\,a). This design was deployed in CVRN~\cite{huang2021cross}. The two encoders utilize the same architecture but the decoder architecture are task-specific. A bottom-up decoder is used for semantic segmentation while a top-down decoder is used for instance segmentation. The design of the network building blocks is further detailed in \S\ref{sec:method_edaps_architecture}. A potential disadvantage of M-Net is the use of separate encoders, which prevents the network from jointly adapting task-shared features from the source to the target domain.

\paragraph{S-Net}

shares the encoder and decoder for both tasks (see Fig.~\ref{fig:architecture_comparison}\,b), which is similar to the architecture of UniDAPS~\cite{zhang2022hierarchical}. As the decoder is shared, it is not possible to have task-specific designs, which can be disadvantageous if different decoder architectures have better domain adaptation properties for semantic and instance segmentation. Also, shared decoders can favor negative transfer of task-specific knowledge.

\paragraph{M-Dec-BU}

shares the encoder but splits the decoders to enable task-specific knowledge. Both decoders use a bottom-up design.

\paragraph{M-Dec-TD (EDAPS)}

shares the encoder but splits the decoders to enable task-specific knowledge. The instance decoder uses a top-down approach in contrast to the bottom-up semantic decoder (see Fig.~\ref{fig:architecture_comparison}\,c). We adopt this design in our EDAPS architecture. In the following section, we motivate and detail the design choices of the EDAPS architecture.

\begin{figure*}[t]
  \centering
   
   \includegraphics[width=1.\linewidth]{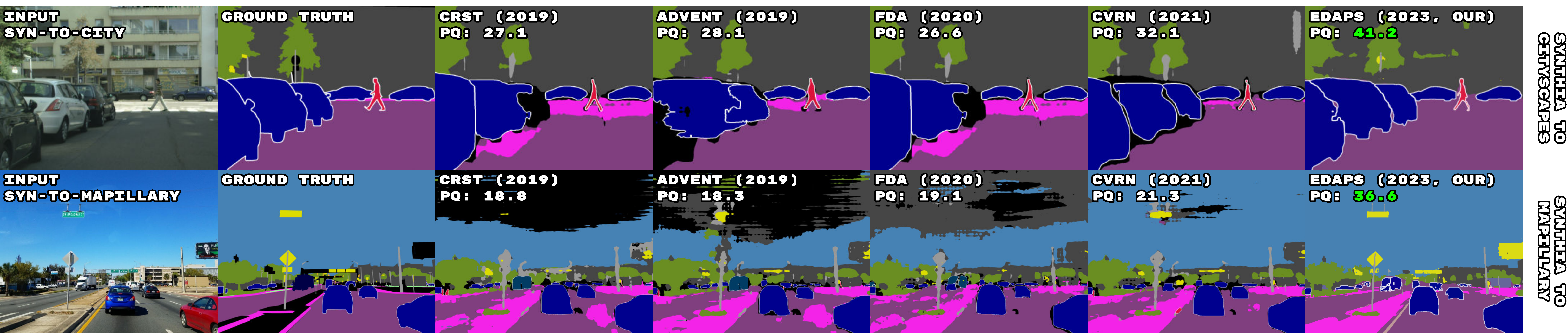}
   \caption{
   Visual comparison of the panoptic segmentation quality of EDAPS (our method) with the prior works CRST \cite{zou2019confidence}, FDA \cite{yang2020fda}, AdvEnt \cite{vu2019advent}, and CVRN \cite{huang2021cross} on the two UDA benchmarks: SYNTHIA $\rightarrow$ Cityscapes (top) and SYNTHIA $\rightarrow$ Mapillary Vistas (bottom).
   }
   \label{fig:qualitatives}
\end{figure*}

\subsection{EDAPS Network Architecture}
\label{sec:method_edaps_architecture}

An overview of the proposed EDAPS architecture is presented in Fig.~\ref{fig:overview}. On the one side, EDAPS utilizes a shared transformer backbone, which has an improved domain-robustness compared to CNNs \cite{naseer2021intriguing,hoyer2021daformer}. Sharing the encoder is advantageous compared the separate networks as it can facilitate the joint adaptation of semantic and instance features from the source to the target domain. On the other side, EDAPS uses task-specific decoders with different design tailored for the specific requirements of domain-adaptive semantic segmentation and domain-adaptive instance segmentation. In that way, EDAPS combines the strength of previous methods (see Fig.~\ref{fig:architecture_comparison}).

\paragraph{Shared Hierarchical Transformer Encoder} 
Transformers have shown domain-robust properties beneficial for UDA~\cite{hoyer2021daformer,xu2021cdtrans,sun2022safe}.
Therefore, EDAPS uses a Mix Transformer (MiT-B5)~\cite{xie2021segformer} as its encoder network, which is tailored for dense predictions by generating coarse (high-resolution) and fine-grained (low-resolution) features at different levels. 
Unlike $16 \time 16$ patches of ViT~\cite{dosovitskiy2020image}, MiT divides an image into relatively smaller patches of shape $4 \times 4$; this facilitates preserving finer details helpful in improving semantic segmentation.
The patches are processed with four transformer blocks into multi-level features at $\{1/4, 1/8, 1/16, 1/32\}$ scales of the original image shapes.
Moreover, an efficient self-attention is used to cope with the computation bottleneck of transformer encoders, i.e., the computational complexity of the self-attention operation is minimized by reducing the input sequence length $(H \times W)$ using a reduction ratio as in \cite{wang2021pyramid}.
Overlapping patch merging is used to downsample feature maps which helps preserve local continuity.
These properties of MiT make it possible to learn robust domain-invariant features for UDA panoptic segmentation using limited computing resources (i.e., on a single GPU).

\paragraph{Bottom-Up Semantic Decoder} 
We adopt a domain-robust context-aware feature fusion decoder~\cite{hoyer2021improving} as our semantic decoder, which has two main benefits over traditional DeepLab decoders~\cite{chen2017deeplab,george2018encoder} or MLP-based decoders~\cite{wang2021pyramid,xie2021segformer,zheng2021rethinking}.
Firstly, it allows exploiting context information alongside local information in the decoder, which is beneficial for domain-robust semantic segmentation~\cite{kamann2021benchmarking}.
Secondly, instead of only using the context information coming from the bottleneck features~\cite{chen2017deeplab,george2018encoder}, it fuses context encoded by features at different levels of the backbone network.
The high-resolution features from the earlier layers encode low-level concepts helpful in better semantic segmentation.
The multi-level context-aware feature fusion is learned using a series of $3 {\times} 3$ depthwise separable convolutions~\cite{chollet2017xception} with different dilation rates.

\paragraph{Top-Down Instance Decoder} 

Based on the findings of the network study in \S\ref{sec:exp}, EDAPS resorts to a proposal-based top-down instance decoder, which exhibits better domain-adaptive properties compared to a bottom-up decoder.

Specifically, we follow Mask R-CNN~\cite{he2017mask} and use a Region Proposal Network (RPN) to predict candidate object bounding boxes. It is trained using a binary cross-entropy (CE) loss for bounding box classification (object or no object), and an $L_1$ loss for box regression.
Given the bounding boxes predicted by RPN, a Region-of-Interest Alignment (RoIAlign) layer is used to extract features from each box. 
The extracted features are used as inputs to the box and mask heads.
The box head is trained to predict the bounding boxes and class labels, and the mask head is trained to predict the class-agnostic binary instance masks.
The box head is trained using CE loss for object classification and $L_1$ loss for box regression over all thing classes.
The mask head predictions are penalized using the binary CE loss.

\paragraph{Bottom-Up Instance Decoder (Baseline)}

For the network studies with a bottom-up instance decoder, we adapt the strong domain-robust decoder from~\cite{hoyer2021improving} from semantic segmentation to instance segmentation by replacing the classification head (for semantic prediction) with two $1 \times 1$ convolutional layers that predict the instance centers and offsets. It is trained using the losses proposed in~\cite{cheng2019panoptic}.

\paragraph{Feature Fusion}

We follow Panoptic-DeepLab~\cite{cheng2019panoptic} to fuse semantic and instance predictions into the final panoptic segmentation upon inference.
Class-agnostic instance segmentation maps are generated by selecting the top-k instance predictions with a detection score above a certain threshold. 
The resulting instance segmentation is fused with the predicted semantic map by a majority-voting rule to generate the final panoptic segmentation.

%% file: text/exp.tex
\section{Experiments}
\label{sec:exp}

\subsection{Implementation Details}

\paragraph{Datasets}
We evaluate EDAPS for the common setting of synthetic-to-real adaptation. As a source dataset, we use Synthia~\cite{ros2016synthia}, which contains 9,400 synthetic images and panoptic labels. As target datasets, we use Cityscapes~\cite{cordts2016cityscapes}  and Mapillary Vistas~\cite{neuhold2017mapillary}.
Cityscapes~\cite{cordts2016cityscapes} consists of 2,975 training and 500 validation images of European street scenes.
Mapillary Vistas~\cite{neuhold2017mapillary} is a large-scale autonomous driving dataset consisting of 18,000 training and 2,000 validation images.

\paragraph{Training}
We follow DAFormer~\cite{hoyer2021daformer} and train EDAPS with AdamW~\cite{loshchilov2018decoupled} for 40k iterations, a batch size of 2, a crop size of 512$\times$512, a learning rate of $6 {\times} 10^{-5}$ with a linear warmup for 1.5k iterations and polynomial decay afterward, and a weight decay of 0.01.
The experiments are conducted on a RTX 2080 Ti GPU with 11 GB memory.
We use the same data augmentation parameters as in DACS~\cite{tranheden2021dacs}.
To encourage the under-represented classes (in the source domain) to be sampled more frequently, we use RCS~\cite{hoyer2021daformer}
and set the RCS temperature to $0.01$.
Following DAFormer~\cite{hoyer2021daformer}, we also use the ImageNet feature distance loss to preserve 
information about certain thing classes encoded in the ImageNet features.
We set this loss weight $\lambda_\mathit{FD}=0.005$.
For the bottom-up instance head, we use the center loss weight $\lambda_\mathrm{heatmap}=10.0$, offset loss weight $\lambda_\mathrm{offs}=0.1$.
For the top-down instance head, we set the following loss weights to $1.0$:
$\lambda_\mathrm{RPN\text{-}cls}$,
$\lambda_\mathrm{RPN\text{-}box}$,
$\lambda_\mathrm{RoI\text{-}cls}$,
$\lambda_\mathrm{RoI\text{-}box}$,
$\lambda_\mathrm{RoI\text{-}mask}$.

\paragraph{Evaluation Metrics}
For evaluating panoptic segmentation, we use the panoptic quality (PQ) metric \cite{kirillov2019panoptic}
that captures performance for all classes (including stuff and things) in an interpretable and unified manner.
PQ can be seen as the multiplication of a segmentation quality (SQ) term and 
a recognition quality (RQ) term, i.e., PQ = SQ $\times$ RQ.
We report PQ for each category.
Mean SQ (mSQ), mean RQ (mSQ), and mean PQ (mPQ) are average scores over all categories.
For evaluating instance and semantic segmentation, we use the mIoU and mAP metrics respectively.

\begingroup
\setlength{\tabcolsep}{4pt} 
\begin{table*}[ht!]
\normalsize
\centering
\caption{
Comparison with state-of-the-art methods on SYNTHIA $\rightarrow$ Cityscapes benchmark for UDA panoptic segmentation. 
For clarity, per class PQs are reported. The results of EDAPS are averaged over 3 random seeds.
}
\footnotesize
\input{tables/edaps_sota_s2c}
\label{table:s2c_pq}
\end{table*}
\endgroup

\begingroup
\setlength{\tabcolsep}{4pt} 
\begin{table*}[ht!]
\normalsize
\centering
\caption{
Comparison with state-of-the-art methods on SYNTHIA $\rightarrow$ Mapillary Vistas benchmark for UDA Panoptic Segmentation.
For clarity, per class PQs are reported. The results of EDAPS are averaged over 3 random seeds.
}
\footnotesize
\input{tables/edaps_sota_s2m}

\label{table:s2m_pq}
\end{table*}
\endgroup

\begin{table}[t]
\normalsize
\centering
\caption{
Comparison on Cityscapes $\rightarrow$ Foggy Cityscapes and Cityscapes $\rightarrow$ Mapillary Vistas.}
\vspace{-0.1cm}
\footnotesize
\setlength{\tabcolsep}{3.5pt}
\input{tables/results_on_new_uda_benchmarks.tex}

\label{tab:results_on_new_uda_benchmarks}
\vspace{-0.1cm}
\end{table}

\input{tables/relative_uda_performance}

\subsection{Comparison with the State of the Art}

First, we compare our EDAPS with the state-of-the-art methods for panoptic segmentation UDA on 
SYNTHIA $\rightarrow$ Cityscapes (Tab.~\ref{table:s2c_pq}), SYNTHIA $\rightarrow$ Mapillary (Tab.~\ref{table:s2m_pq}),
Cityscapes $\rightarrow$ Foggy Cityscapes, and Cityscapes $\rightarrow$ Mapillary Vistas (Tab.~\ref{tab:results_on_new_uda_benchmarks}).
It can be seen that EDAPS consistently outperforms previous methods with a large margin in all aggregated metrics (mSQ, mRQ, mPQ). Specifically, EDAPS improves the mPQ from 34.2 to 41.2 on SYNTHIA $\rightarrow$ Cityscapes and from 21.3 to 36.6 SYNTHIA $\rightarrow$ Mapillary, which is a respective improvement of remarkable 20\% and 72\% over previous works. 
Tab.~\ref{table:s2c_pq} shows that EDAPS significantly improves both the recognition (+9.3 mRQ) as well as the segmentation quality (+5.8 mSQ) over previous SOTA. Both can be also observed qualitatively in Fig.~\ref{fig:qualitatives}.
EDAPS particularly improves the performance of classes that previous methods struggled with, such as wall, pole, traffic light, traffic sign, person, rider on Cityscapes (Tab.~\ref{table:s2c_pq}), and road, sidewalk, building, wall, pole, traffic light, traffic sign, vegetation, sky, bus on Mapillary (Tab.~\ref{table:s2m_pq}).
Additional benchmarks for other domain gaps are presented in Tab.~\ref{tab:results_on_new_uda_benchmarks}.
EDAPS also improves the SOTA on Cityscapes {$\rightarrow$} Foggy Cityscapes and Cityscapes {$\rightarrow$} Mapillary Vistas with respective improvements of 47\% and 23\%
over UniDAPS and CVRN with fewer parameters and  faster inference speed (Tab.~\ref{tab:efficiency}).
As the different panoptic UDA methods resort to different network architectures, it has to be considered that they have a varying inherent capability to learn panoptic segmentation. Therefore, we compare both the UDA performance and the supervised performance (as indicators of the capability of a network architecture in Tab.~\ref{tab:relative_uda}. Further, we normalize the UDA performance by the supervised performance ($\text{mPQ}_\text{Rel}$) for a more fair comparison. 
It can be seen that even though UniDAPS uses a more powerful network than CVRN, as indicated by the higher supervised performance, this does not translate well to UDA performance. Therefore, UniDAPS has a significantly lower relative UDA performance than CVRN. The lower relative performance of UniDAPS could be caused by the additional transformer encoder-decoder architecture added to the backbone, which is not pre-trained on ImageNet and might overfit to the source domain more easily.
In contrast, our method can increase the relative UDA performance alongside with the best UDA and the best supervised performance. This means that EDAPS can effectively narrow the domain gap to supervised learning.

\begingroup
\setlength{\tabcolsep}{3.5pt} 
\begin{table}
\centering
\caption{
Efficiency comparison on an RTX 2080 Ti.
}
\footnotesize
\input{tables/efficiency}
\label{tab:efficiency}
\end{table}
\endgroup

We compare EDAPS with previous works from an efficiency viewpoint in Tab.~\ref{tab:efficiency}. EDAPS with a MiT-B5 backbone (default) increases the inference speed by a factor of 16 compared to CVRN, demonstrating the efficiency of EDAPS. Compared to UniDAPS, EDAPS w/ MiT-B5 reaches 80\% of its inference speed, which is due to the additional instance decoder of EDAPS. However, given the large gains in the quality of the adapted panoptic segmentation, this is an acceptable trade-off. Even with a smaller MiT-B2 backbone, EDAPS outperforms UniDAPS by +4.1 mPQ on SYNTHIA $\to$ Cityscapes while requiring fewer parameters
and with faster inference time (9.0 vs. 7.2 fps).

\subsection{Study of Panoptic Architectures for UDA}
\label{sec:network_study}
\begingroup
\setlength{\tabcolsep}{3.5pt} 
\begin{table}
\centering
\caption{
Network topology study as detailed in \S\ref{sec:methods_principal_architectures}. Mean and standard deviation are computed over 3 random seeds.
}
\footnotesize
\input{tables/network_study}

\label{tab:network_study}
\end{table}
\endgroup

\begingroup
\setlength{\tabcolsep}{1pt}
\renewcommand{\arraystretch}{1.3}
\begin{table}
\centering
\caption{Class-wise PQ comparison of EDAPS and M-Net.}
\footnotesize
\resizebox{\linewidth}{!}{
\input{tables/class_study_2}
}
\label{tab:class_wise_pq_comparison}
\end{table}
\endgroup

To gain insights into the influence of the network architecture on the UDA performance, we compare the four principal architectures from \S\ref{sec:methods_principal_architectures}.
For all network variants, we use the enhanced network components of EDAPS, i.e., the transformer encoder and the domain-robust decoder as well as the same enhanced UDA strategy for a fair comparison.

Tab.~\ref{tab:network_study} shows that S-Net achieves the lowest mPQ of 34.0. In contrast, M-Net achieves 38.1 mPQ, which is +4.1 mPQ better. This demonstrates the importance of task-specific networks (components) to learning task-specific knowledge in the UDA setting. Still, sharing features across tasks in a common encoder for UDA is useful, as shown by M-Dec-BU, which gains a performance improvement of +0.9 mPQ over M-Net. However, the symmetric architecture with bottom-up decoders for both semantics and instances is sub-optimal, as can be seen when comparing M-Dec-BU with M-Dec-TD. With a top-down decoder for instance segmentation, a significant improvement of +2.2 mPQ can be achieved over M-Dec-BU. In particular, the instance mAP improves by +16.8. This shows that the top-down approach is more domain-robust for instance segmentation than the bottom-up approach.

The network study in Tab.~\ref{tab:network_study} also allows a fair comparison of EDAPS and CVRN as M-Net uses the
CVRN architecture with the enhanced components of EDAPS, i.e., Transformer encoder, Mean Teacher, RCS, and FD. Therefore, M-Net has a higher 38.1 mPQ compared to the original CVRN 32.1 mPQ. Still, EDAPS outperforms M-Net by +3.1 mPQ. These improvements come from the joint training of EDAPS for semantic and instance segmentation (compared to disjoint M-Net training), allowing the shared encoder to learn better domain-robust instance and semantic features. In particular, EDAPS improves the instance mAP by +11.1 over M-Net demonstrating that particularly the instance head benefits from the shared features. The class-wise comparison in Tab.~\ref{tab:class_wise_pq_comparison} reveals that EDAPS mostly benefits difficult thing-classes such as rider or motorbike, showing that shared features are particularly important for instances that are difficult to adapt.

\subsection{UDA Ablation Study}

\begingroup
\setlength{\tabcolsep}{3.5pt} 
\begin{table}
\centering
\caption{
Ablation study of the UDA strategies Self-Training (Self-Tr.), Mean Teacher (MT), ImageNet Feature Distance (FD), and Rare Class Sampling (RCS). Mean and standard deviation are provided over 3 random seeds.
}
\footnotesize
\input{tables/uda_ablation}

\label{tab:uda_ablation}
\end{table}
\endgroup

To better understand the influence of the UDA components that we newly introduced to panoptic UDA, we ablate them in Tab.~\ref{tab:uda_ablation}. While the source-only model performs at 21.6 mPQ, the UDA baseline with basic self-training achieves a performance of 37.5 mPQ. This is already higher than previous state-of-the-art methods, emphasizing the strength of the proposed EDAPS architecture. The self-training can be improved by +1.4 mPQ using an EMA teacher for pseudo-label generation. ImageNet Thing-Class Feature Distance (FD) further increases performance by +0.8 mPQ. When further integrating Rare Class Sampling (RCS), the performance gains another +1.5 mPQ. This shows that the EMA teacher, FD and RCS are valuable components for panoptic UDA.
Empirically, we found that additional instance pseudo-labels do not improve the performance of EDAPS. We assume that semantic self-training is sufficient to bridge the domain gap at the level of the shared encoder features. For simplicity, we therefore only use semantic pseudo-labels.

%% file: tables/edaps_sota_s2c.tex
\begin{tabular}{l @{\quad} cccccccccccccccc @{\quad} c @{\quad} c @{\quad} c}
\toprule 
UDA Method & \rots{road} & \rots{sidewalk\quads\quads} & \rots{building} & \rots{wall} & \rots{fence} & \rots{pole} & \rots{light} & \rots{sign} & \rots{veg} & \rots{sky} & \rots{person} & \rots{rider} & \rots{car} & \rots{bus} & \rots{m.bike} & \rots{bike} & mSQ & mRQ & mPQ \\

\midrule
FDA \cite{yang2020fda}                  & 79.0      & 22.0      & 61.8      & 1.1  & 0.0 & 5.6  & 5.5   & 9.5  & 51.6 & 70.7 & 23.4 & 16.3 & 34.1 & 31.0 & 5.2  & 8.8  & 65.0 & 35.5 & 26.6 \\
CRST \cite{zou2019confidence}           & 75.4      & 19.0      & 70.8      & 1.4  & 0.0 & 7.3  & 0.0   & 5.2  & 74.1 & 69.2 & 23.7 & 19.9 & 33.4 & 26.6 & 2.4  & 4.8  & 60.3 & 35.6 & 27.1 \\
AdvEnt \cite{vu2019advent}              & 87.1      & 32.4      & 69.7      & 1.1  & 0.0 & 3.8  & 0.7   & 2.3  & 71.7 & 72.0 & 28.2 & 17.7 & 31.0 & 21.1 & 6.3  & 4.9  & 65.6 & 36.3 & 28.1 \\
CVRN \cite{huang2021cross}              & 86.6  & 33.8      & 74.6      & 3.4  & 0.0 & 10.0 & 5.7   & 13.5 & 80.3 & 76.3 & 26.0 & 18.0 & 34.1 & 37.4 & 7.3  & 6.2  & 66.6 & 40.9 & 32.1 \\
UniDAPS \cite{zhang2022hierarchical}    & 73.7      & 26.5      & 71.9      & 1.0 & 0.0 & 7.6 & 9.9 & 12.4 & 81.4 & 77.4 & 27.4 & 23.1 & \B{47.0} & \B{40.9} & 12.6 & 15.4 & 64.7 & 42.2 & 33.0 \\
\;\rotatebox[origin=c]{180}{$\Lsh$} w/ CVRN Net \cite{zhang2022hierarchical} & \B{87.7} & 34.0 & 73.2 & 1.3 & 0.0 & 8.1 & 9.9 & 6.7 & 78.2 & 74.0 & 37.6 & 25.3 & 40.7 & 37.4 & 15.0 & \B{18.8} & 66.9 & 44.3 & 34.2\\
\midrule
EDAPS (Ours)                            & 77.5      & \B{36.9}  & \B{80.1} & \B{17.2} & \B{1.8} & \B{29.2} & \B{33.5} & \B{40.9} & \B{82.6} & \B{80.4} & \B{43.5} & \B{33.8} & 45.6 & 35.6 & \B{18.0} & 2.8 & \B{72.7} & \B{53.6} & \B{41.2} \\

\bottomrule
\end{tabular}

%% file: tables/edaps_sota_s2m.tex
\begin{tabular}{l @{\quad} cccccccccccccccc @{\quad} c @{\quad} c @{\quad} c}
\toprule 
UDA Method & \rots{road} & \rots{sidewalk\quads\quads} & \rots{building} & \rots{wall} & \rots{fence} & \rots{pole} & \rots{light} & \rots{sign} & \rots{veg} & \rots{sky} & \rots{person} & \rots{rider} & \rots{car} & \rots{bus} & \rots{m.bike} & \rots{bike} & mSQ & mRQ & mPQ \\

\midrule
FDA \cite{yang2020fda}          & 44.1 & 7.1 & 26.6 & 1.3 & 0.0 & 3.2 & 0.2 & 5.5 & 45.2 & 61.3 & 30.1 & 13.9 & 39.4 & 12.1 & 8.5 & 7.0 & 63.8 & 26.1 & 19.1 \\
CRST \cite{zou2019confidence}   & 36.0 & 6.4 & 29.1 & 0.2 & 0.0 & 2.8 & 0.5 & 4.6 & 47.7 & 68.9 & 28.3 & 13.0 & 42.4 & 13.6 & 5.1 & 2.0 & 63.9 & 25.2 & 18.8 \\
AdvEnt \cite{vu2019advent}      & 27.7 & 6.1 & 28.1 & 0.3 & 0.0 & 3.4 & 1.6 & 5.2 & 48.1 & 66.5 & 28.4 & 13.4 & 40.5 & 14.6 & 5.2 & 3.3 & 63.6 & 24.7 & 18.3 \\
CVRN \cite{huang2021cross}      & 33.4 & 7.4 & 32.9 & 1.6 & 0.0 & 4.3 & 0.4 & 6.5 & 50.8 & 76.8 & 30.6 & 15.2 & 44.8 & 18.8 & 7.9 & \B{9.5} & 65.3 & 28.1 & 21.3 \\
\midrule
EDAPS (Ours)                    & \B{77.5} & \B{25.3} & \B{59.9} & \B{14.9} & 0 & \B{27.5} & \B{33.1} & \B{37.1} & \B{72.6} & \B{92.2} & \B{32.9} & \B{16.4} & \B{47.5} & \B{31.4} & \B{13.9} & 3.7 & \B{71.7} & \B{46.1} & \B{36.6} \\

\bottomrule
\end{tabular}

%% file: tables/results_on_new_uda_benchmarks.tex
\begin{tabular}{lcc c lcc}
\toprule
 \multicolumn{3}{c}{Cityscapes $\rightarrow$ Foggy Cityscapes}           &  &  \multicolumn{3}{c}{Cityscapes $\rightarrow$ Mapillary Vistas} \\
\midrule
UDA Method                              & \#Params      &  mPQ            &  &   UDA Method                  &   \#Params        &  mPQ \\
  \midrule
UniDAPS~\cite{zhang2022hierarchical}    & 58.7 M        & 37.6            &  & CVRN~\cite{huang2021cross}    & 185.5 M           & 33.5  \\
EDAPS-MiTB2	                            & 47.6 M        & 55.1            &  & EDAPS-MiTB5                   & 104.9 M           & 41.2   \\
\bottomrule
\end{tabular}

%% file: tables/relative_uda_performance.tex
\begin{table}
\caption{Comparison of the mPQ for source-only training, UDA (SYNTHIA $\rightarrow$ Cityscapes), and supervised oracle training (Cityscapes) along with the relative UDA performance $\text{mPQ}_\text{Rel} {=} \text{mPQ}_\text{UDA} / \text{mPQ}_\text{Sup}$.
}
\label{tab:relative_uda}
\centering
\footnotesize
\begin{tabular}{lcccc}
\toprule
UDA Method  & mPQ$_\text{Src-Only}$ & mPQ$_\text{UDA}$ & mPQ$_\text{Sup}$ & mPQ$_\text{Rel}$ \\
\midrule
CVRN~\cite{huang2021cross}        & 20.1 & 32.1    & 47.7    & 67.3\%   \\
UniDAPS~\cite{zhang2022hierarchical}     & 18.3 & 33.0    & 51.9    & 63.6\%   \\
EDAPS (Ours) & \textbf{21.6} & \textbf{41.2}    & \textbf{56.6}    & \textbf{72.7\%}   \\
\bottomrule
\end{tabular}
\end{table}

%% file: tables/efficiency.tex
\begin{tabular}{l @{\quad} ccccc}

\toprule 

Network architecture & \#Parameters & Inference Speed & mPQ$_{\text{SYN}\to\text{CS}}$ \\
\midrule
CVRN~\cite{huang2021cross}                  & 185.5 M   & 0.36 fps & 32.1\\
UniDAPS~\cite{zhang2022hierarchical}        & 58.7 M    & 7.24 fps & 33.0\\
EDAPS w/ MiT-B2                             & 47.6 M   & 9.03 fps & 37.1\\
EDAPS w/ MiT-B3                             & 67.5 M   & 7.65 fps & 38.9\\
EDAPS w/ MiT-B5                             & 104.9 M   & 5.84 fps & 41.2\\
\bottomrule
\end{tabular}

%% file: tables/network_study.tex
\setlength\tabcolsep{1px}
\begin{tabular}{l @{\quad} ccccc}

\toprule 

Network
& mAP & mIoU & mSQ & mRQ & mPQ \\

\midrule

S-Net
& 7.0\spm{0.4} 
& 57.8\spm{1.1}  
& 72.0\spm{0.4} 
& 43.8\spm{0.8} 
& 34.0\spm{0.7} 
\\

M-Net
& 23.3\spm{0.3} 
& 56.6\spm{0.6} 
& 71.7\spm{0.3} 
& 50.0\spm{0.5} 
& 38.1\spm{0.3} 
\\

M-Dec-BU      
& 17.6\spm{1.5} 
& 60.3\spm{0.6} 
& 73.9\spm{0.4} 
& 49.9\spm{0.3} 
& 39.0\spm{0.2} 
\\

M-Dec-TD (EDAPS)
& 34.4\spm{0.5} 
& 57.5\spm{0.0} 
& 72.7\spm{0.2} 
& 53.6\spm{0.5} 
& 41.2\spm{0.4} 
\\
\bottomrule
\end{tabular}

%% file: tables/class_study_2.tex
\begin{tabular}{lrrrrrrrrrrrrrrrr}
\toprule 

Network & \rots{road} & \rots{sidew.\quads\quads} & \rots{build.} & \rots{wall} & \rots{fence} & \rots{pole} & \rots{light} & \rots{sign} & \rots{veg.} & \rots{sky} & \rots{pers.} & \rots{rider} & \rots{car} & \rots{bus} & \rots{m.bike} & \rots{bike} \\

\midrule
M-Net     & 75.2 & 37.5 & 80.4 & 15.7 & 2.0  & 27.1 & 36.2 & 38.0 & 79.8 & 80.0 & 38.2 & 25.8 & 38.9 & 28.0 & 6.1  & 0.3 \\
EDAPS                 & 77.5      & 36.9  & 80.1 & 17.2 & 1.8 & 29.2 & 33.5 & 40.9 & 82.6 & 80.4 & 43.5 & 33.8 & 45.6 & 35.6 & 18.0 & 2.8 \\
\hline
Gain & +2.3 & -0.6 & -0.3 & +1.5 & -0.2 & +2.1 & -2.7 & +2.9 & +2.8 & +0.4 & \textbf{+5.3} & \textbf{+8.0} & \textbf{+6.7} & \textbf{+7.6} & \textbf{+11.9} & +2.5 \\

\bottomrule
\end{tabular}

%% file: tables/uda_ablation.tex
\setlength\tabcolsep{1.8px}
\begin{tabular}{cccc @{\quad} ccccc}
\toprule 
Self-Tr. & MT & FD & RCS & mAP & mIoU & mSQ & mRQ & mPQ \\
\midrule

& 
& 
& 
& 22.0\spm{1.0}
& 33.0\spm{1.8}
& 60.9\spm{2.6} 
& 29.3\spm{1.7} 
& 21.6\spm{1.2} 
\\

\checkmark 
& 
& 
& 
& 34.7\spm{0.9} 
& 54.0\spm{0.1} 
& 71.5\spm{0.4} 
& 49.1\spm{0.4} 
& 37.5\spm{0.2}
\\

\checkmark 
& 
\checkmark
& 
& 
& 35.7\spm{1.7} 
& 56.3\spm{0.9} 
& 69.3\spm{1.6} 
& 50.7\spm{1.5} 
& 38.9\spm{1.2} 
\\

\checkmark 
& 
\checkmark
& 
\checkmark
& 
& 35.2\spm{0.9} 
& 56.7\spm{0.9} 
& 70.7\spm{1.8} 
& 52.1\spm{1.1} 
& 39.7\spm{0.9} 
\\

\checkmark 
& 
\checkmark
& 
\checkmark
& 
\checkmark
& 34.4\spm{0.5} 
& 57.5\spm{0.0} 
& 72.7\spm{0.2} 
& 53.6\spm{0.5} 
& 41.2\spm{0.4} 
\\

\bottomrule
\end{tabular}

%% file: text/conclusion.tex
\section{Conclusions}

Previous approaches to domain-adaptive panoptic segmentation either follow inefficient and expensive adaptation techniques, or their network architecture and training strategies are influenced by supervised learning.
In this work, we addressed these issues by carefully selecting the network design and training schemes tailored for UDA and proposed EDAPS, an efficient domain-adaptive panoptic segmentation network that 
surpasses prior art by a large margin. 
Furthermore, we provided a detailed analysis of the various design choices for enhancing the panoptic UDA performance.
When compared with previous methods on UDA panoptic segmentation, EDAPS shows significant performance gains of 20\% on SYNTHIA-to-Cityscapes and even 72\% on the more challenging SYNTHIA-to-Mapillary Vistas.
We believe that EDAPS as a network architecture will facilitate benchmarking future UDA strategies in panoptic segmentation, making domain-adaptive panoptic segmentation more usable in practice.

%% file: text/sup_mat.tex
\section{Overview}

This supplementary material provides a more detailed analysis of the experiments presented in the paper.
In particular,
Sec.~\ref{sec:further_impl_details} provides further implementation details,
Sec.~\ref{sec:real_to_real_uda_benchmarks} presents a detailed class-wise performance comparison on additional UDA benchmarks, Sec.~\ref{sec:adv_vs_self_training} highlights the benefits of self-training over adversarial training for UDA panoptic segmentation,
Sec.~\ref{sec:ablation_study} presents an ablation study showing the significance of different instance losses on the adaptation process,
Sec.~\ref{sec:qualitative_analysis} analyzes additional qualitative example predictions,
and Sec.~\ref{sec:mdecbu-analysis} offers a visual comparison of the predictions made by EDAPS and M-Dec-BU.

\section{Further Implementation Details}
\label{sec:further_impl_details}
EDAPS is implemented in PyTorch \cite{paszke2017automatic} based on the DAFormer framework~\cite{hoyer2021daformer}.
The source code is available at \url{https://github.com/susaha/edaps} to ensure easy reproducibility and promote research in domain-adaptive panoptic segmentation.
We follow CVRN \cite{huang2021cross} and consider $11$ stuff-classes and $8$ thing-classes.
The stuff classes are road, sidewalk, building, wall, fence, pole, traffic light, traffic sign, vegetation, terrain, and sky;
the thing classes are person, rider, car, truck, bus, train, motorcycle, and bicycle. 

We use a threshold of $0.95$ to select the top-k binary masks predicted by the EDAPS instance head.
We use these top-k predicted masks to generate the class-agnostic instance segmentation maps, which are then fused with the predicted semantic segmentation maps by a majority-voting rule.

For the Foggy Cityscapes dataset \cite{sakaridis2018model}, we use the attenuation coefficient $\beta = 0.02$.
It specifies the meteorological optical range (MOR) or the visibility, and it is measured in inverse meters. $\beta = 0.02$ corresponds to a MOR of $150$m and represents a considerable domain gap to clear weather scenes (see Fig.~\ref{fig:c2fcs}).

\textbf{M-Dec-BU (Baseline).}
Since the bottom-up instance decoder (used in the M-Dec-BU) does not directly predict instance masks,
a post-processing step is required to generate the class-agnostic instance segmentation maps 
from the predicted center and offset heatmaps. 
The post-processing step includes selecting the top-k instance centers and
grouping pixels based on these selected centers.
We pick the top-k predicted centers by first applying a hard thresholding to filter out the low-confident center predictions following a 2D max pooling on the predicted center heatmap.
In all our experiments, 
we set the threshold to $0.1$,
max-pooling kernel size to $7 \times 7$,
and $k = 200$ as in \cite{cheng2019panoptic}.

Once the top-k instance centers are selected, 
we assign each pixel an instance id based on the predicted offset heatmap. 
More specifically, the instance id for a pixel is the index of the closest instance center 
after moving the pixel location by the offset.
We filter out the stuff pixels based on the predicted semantic segmentation.
Once the instance ids are computed, 
we generate the class-agnostic instance segmentation maps and 
fuse them with the predicted semantic segmentation maps by a majority-voting rule \cite{cheng2019panoptic}.

\begin{table*}[ht!]
\setlength{\tabcolsep}{3pt} 
\normalsize
\centering
\caption{
Comparison with state-of-the-art methods on Cityscapes $\rightarrow$ Foggy Cityscapes benchmark for UDA Panoptic Segmentation.
For clarity, per class PQs are reported. The results of EDAPS are averaged over 3 random seeds.
}
\footnotesize
\input{tables/edaps_sota_c2fc}

\label{table:edaps_sota_c2fc}
\end{table*}

\begin{table*}[ht!]
\setlength{\tabcolsep}{3pt} 
\normalsize
\centering
\caption{
Comparison with state-of-the-art methods on Cityscapes $\rightarrow$ Mapillary Vistas benchmark for UDA Panoptic Segmentation.
For clarity, per class PQs are reported. The results of EDAPS are averaged over 3 random seeds.
}
\footnotesize
\input{tables/edaps_sota_c2m}
\label{table:edaps_sota_c2m}
\end{table*}

\begin{table}[t]
\normalsize
\centering
\caption{
Performance comparison between adversarial and self training based models
(SYNTHIA {$\rightarrow$} Cityscapes).
}
\vspace{-0.1cm}
\resizebox{\linewidth}{!}{
\input{tables/adversarial_train_analysis.tex}

}
\label{tab:adversarial_train_analysis}
\vspace{-0.1cm}
\end{table}

\begingroup
\setlength{\tabcolsep}{4.5pt} 
\renewcommand{\arraystretch}{1.3}
\begin{table*}[ht!]
\centering
\caption{
EDAPS instance head losses ablation study on the SYNTHIA $\rightarrow$ Cityscapes UDA panoptic benchmark.
The results of the trained models are averaged over 3 random seeds.
}
\footnotesize
\input{tables/mask-rcnn-ablation}
\label{tab:mask_rcnn_ablation}
\end{table*}
\endgroup

\section{Comparison on Additional Benchmarks}
\label{sec:real_to_real_uda_benchmarks}
In this section, we report UDA performance on additional clear-to-foggy and real-to-real  UDA benchmarks.
Tab.~\ref{table:edaps_sota_c2fc} and Tab.~\ref{table:edaps_sota_c2m} present comparisons with state-of-the-art methods on 
Cityscapes $\rightarrow$ Foggy Cityscapes 
and Cityscapes $\rightarrow$ Mapillary Vistas 
benchmarks.
We report a detailed class-wise PQ comparison to gain a better insight into the performance analysis.
EDAPS shows significant performance gains for most of the \emph{thing} and \emph{stuff} classes.
Most importantly, 
EDAPS significantly improves the mean recognition quality (mRQ) on both clear-to-foggy and real-to-real benchmarks with a respective percentage gain of 42\% (Tab.~\ref{table:edaps_sota_c2fc})
and 25\% (Tab.~\ref{table:edaps_sota_c2m}).

\section{Adversarial- vs. Self-Training}
\label{sec:adv_vs_self_training}
We chose self-training over adversarial training because it is the predominant SOTA approach in UDA semantic segmentation. 
Further, adversarial training is rather unstable, which makes our architecture study more difficult.
To provide a more comprehensive picture, we additionally train EDAPS with adversarial training \cite{tsai2018learning}
in Tab.~\ref{tab:adversarial_train_analysis}.
Consistent with UDA semantic segmentation, self-training achieves better results for UDA panoptic segmentation.

\section{Ablation Study of Instance Losses}
\label{sec:ablation_study}
EDAPS uses a top-down instance decoder, which is trained using $5$ losses. Even though the effect of these losses is well explored for supervised panoptic segmentation, the influence of the different losses on UDA panoptic segmentation has not been studied so far. Therefore, we present a detailed ablation study analyzing the effect of each instance loss on the domain-adaptive panoptic performance (mPQ). Furthermore, we provide the domain-adaptive instance segmentation performance (mAP), which helps to understand the significance of each instance loss towards the adaptation process for instance segmentation.

We ablate all $5$ instance losses, including the losses in the RPN and RoI heads.
There are $2$ losses in the RPN head, RPN bounding-box classification and regression losses ($\mathcal{L}_\mathrm{RPN\text{-}cls}$, $\mathcal{L}_\mathrm{RPN\text{-}box}$),
and $3$ losses in the RoI head, RoI bounding-box classification, regression, and RoI mask classification losses ($\mathcal{L}_\mathrm{RoI\text{-}cls}$, 
$\mathcal{L}_\mathrm{RoI\text{-}box}$, $\mathcal{L}_\mathrm{RoI\text{-}mask}$).
For this ablation, we train $8$ models with different combinations of the instance losses on the SYNTHIA $\rightarrow$ Cityscapes benchmark.
The models are trained following the same setup as EDAPS.

The results of the ablation study in Table \ref{tab:mask_rcnn_ablation} provide interesting observations:
Without RPN losses, the mPQ decreases from $41.2$ to $30.8$.
At a closer look, we note that instance segmentation (mAP) and recognition quality (mRQ) are adversely affected the most.
That implies, in the absence of good quality region proposals, the network struggles to generate correct instance segmentation masks, and there is an increase in false detections (false positives and false negatives).
Besides, the RPN box regression loss contributes more to the overall performance improvement than the RPN box classification loss. 

In the absence of the RoI head's box classification and regression losses,
the model shows the lowest mPQ, mAP, mRQ, and mSQ of $29.5$, $0.5$, $38.4$, $45.5$, respectively.
It implies that the RoI-pooled features play a vital role; the instance head trained without losses on the RoI features struggles to achieve high-quality instance segmentation.
Interestingly, the RoI head's box classification loss contributes more to the overall performance gain than the box regression loss.
Since the RPN box regression loss already helps the network to learn better instance bounding boxes, even if the RoI head box regression loss is turned off, it achieves an mPQ of $38.2$, which is already better than the $32.1$ mPQ of the prior work CVRN~\cite{huang2021cross}.
However, it is crucial for the RoI head to learn the correct box label classification; since the RPN box classification loss is only responsible for providing correct binary labels (object vs. no-object) for the region proposals, the RoI box classification loss helps the instance head to learn correct instance class labels (i.e., the $8$ thing object classes) for the RoI-predicted boxes.
Finally, in the absence of the RoI mask classification loss, the mPQ goes down from $41.2$ to  $32.7$, which shows that it is crucial for the network to learn the correct binary instance masks to achieve better panoptic segmentation quality.

\section{Qualitative Analysis}
\label{sec:qualitative_analysis}
In this section, we provide additional qualitative prediction results for a visual comparison of the proposed EDAPS and the prior art CVRN \cite{huang2021cross}. 
The visualizations for models trained on SYNTHIA $\rightarrow$ Cityscapes are presented in Fig. \ref{fig:pred_person}-\ref{fig:pred_miscellaneous}.
The major improvements come from better panoptic segmentation of the thing classes \emph{person} (Fig. \ref{fig:pred_person}), \emph{rider}  (Fig. \ref{fig:pred_rider_motorbike}), 
\emph{car} (Fig. \ref{fig:pred_car}); and stuff classes \emph{traffic light}, \emph{traffic sign}, \emph{pole} (Fig. \ref{fig:pred_rider_motorbike}, \ref{fig:pred_car}, and \ref{fig:pred_miscellaneous})
across different object scales, appearance, and viewing angles. 
In general, EDAPS can better delineate object boundaries, resulting in better-quality pixel-level panoptic segmentation.
Note that the detected object shapes (e.g., \emph{person}, \emph{rider}, \emph{car}) predicted by the EDAPS resemble more real-world object shapes when compared to CVRN \cite{huang2021cross}.
Thanks to the domain-robust Mix Transformer (MiT-B5)~\cite{xie2021segformer} backbone, 
EDAPS can learn a richer set of domain-invariant semantic and instance features helpful in better segmentation of fine structures.
EDAPS can better segment the occluded object instances in a crowded scene such as 
\emph{person} (Fig. \ref{fig:pred_person} row $1$-$5$), 
\emph{rider} (Fig. \ref{fig:pred_rider_motorbike} row $1$),
\emph{car} (Fig. \ref{fig:pred_car} row $1$-$8$).
Moreover, the \emph{person} segments predicted by EDAPS preserve finer details of the human body even when instances are occluded.
Similar observations can be made for the \emph{rider} and \emph{car} classes.
For large object instances (such as \emph{bus}), EDAPS can segment out the entire object, whereas CVRN fails to do so (Fig. \ref{fig:pred_rider_motorbike} row $8$, Fig. \ref{fig:pred_miscellaneous} row $1$).
EDAPS can provide better segmentation for the \emph{traffic light} 
(Fig. \ref{fig:pred_rider_motorbike}  row $1$, $8$; Fig. \ref{fig:pred_miscellaneous} row $3$, $4$), and 
\emph{traffic sign} 
(Fig. \ref{fig:pred_rider_motorbike}  row $4$, $8$; Fig. \ref{fig:pred_miscellaneous} row $1$, $4$, $5$).

In addition, we show visual qualitative results on SYNTHIA $\rightarrow$ Mapillary Vistas UDA panoptic benchmark 
(Fig. \ref{fig:pred_set8}-\ref{fig:pred_set10}).
EDAPS segments better the \emph{pole} instance (Fig. \ref{fig:pred_set8} row $5$).
In Fig. \ref{fig:pred_set9} and \ref{fig:pred_set10}, we present a visual comparison 
with the Source-Only model.
It can be observed that the Source-Only model struggles to learn the correct class labels and instance masks, whereas EDAPS successfully bridges the domain gap by learning the correct semantics and instances.
EDAPS produces better panoptic segmentation for 
the \emph{bus} (Fig. \ref{fig:pred_set9} row 1, Fig. \ref{fig:pred_set10} row 1, 2, 3),
\emph{rider} (Fig. \ref{fig:pred_set9} row 2, 3, Fig. \ref{fig:pred_set10} row 5),
\emph{motorbike} (Fig. \ref{fig:pred_set9} row 3, Fig. \ref{fig:pred_set10} row 6),
\emph{car} (Fig. \ref{fig:pred_set10} row 3, 4),
\emph{traffic sign} (Fig. \ref{fig:pred_set9} row 6, Fig. \ref{fig:pred_set10} row 1).
Finally, the visual predictions on Cityscapes $\rightarrow$ Foggy Cityscapes are shown in Fig. \ref{fig:c2fcs}.
\\

\section{Visual Comparison: EDAPS vs. M-Dec-BU}
\label{sec:mdecbu-analysis}
This section offers a visual comparison of the predictions made by EDAPS and M-Dec-BU on the SYNTHIA $\rightarrow$ Cityscapes benchmark, as depicted in Fig. \ref{fig:edaps_vs_mdecbu_01}.
We observed that the M-Dec-BU baseline model tends to segment objects (like 
pedestrians, cars, buses, and riders) into smaller parts than necessary (i.e., over-segmentation).
Notice that the pedestrian, car, and bus instances in Figs.~\ref{fig:edaps_vs_mdecbu_01}~(a-d) are over-segmented.
This over-segmentation problem is more prominent in scenes  with large and occluded objects.

The M-Dec-BU model adopts a bottom-up approach for instance segmentation \cite{cheng2019panoptic}.
Unlike top-down methods \cite{he2017mask}, M-Dec-BU's instance head does not directly predict instance segmentation masks.
Rather, it predicts instance centers and offsets.
An additional post-processing step is required to generate the class-agnostic instance segmentation masks 
from these predicted centers and offsets.
We found that the center predictions are not sufficiently robust under a domain shift (even with domain adaptation) to support reliable post-processing on the target domain which leads to an over-segmentation problem as discussed above.
In contrast, we noticed that EDAPS's top-down instance segmentation head 
predicts highly generalizable instance masks on the target domain
resulting an improved instance segmentation performance (mAP 34.4\%) as
compared to 17.6\% mAP of M-Dec-BU.

\begin{figure*}[h!]
\centering
\input{preds/prediction_head}
\input{preds/city_vis_18} 
\input{preds/city_vis_16} 
\input{preds/city_vis_17} 
\input{preds/city_vis_12} 
\input{preds/city_vis_10} 
\input{preds/city_vis_7} 
\input{preds/city_vis_14} 
\input{preds/city_vis_5} 
\input{preds/city_vis_6} 
\input{preds/palette}
\vspace{-0.15cm}
\caption{Example predictions showing better panoptic segmentation for \emph{person} on SYNTHIA $\rightarrow$ Cityscapes.}
\label{fig:pred_person}
\end{figure*}  

\begin{figure*}[h!]
\centering
\input{preds/prediction_head}
\input{preds/city_vis_24} 
\input{preds/city_vis_19} 
\input{preds/city_vis_8} 
\input{preds/city_vis_31} 
\input{preds/city_vis_15} 
\input{preds/city_vis_25} 
\input{preds/city_vis_26} 
\input{preds/city_vis_27} 
\input{preds/palette}
\vspace{-0.15cm}
\caption{Example predictions showing better panoptic segmentation for \emph{rider}, \emph{motorbike}, \emph{bus}, and \emph{sign} classes on SYNTHIA $\rightarrow$ Cityscapes.}
\label{fig:pred_rider_motorbike}
\end{figure*}

\begin{figure*}[h!]
\centering
\input{preds/prediction_head}
\input{preds/city_vis_20} 
\input{preds/city_vis_23} 
\input{preds/city_vis_9} 
\input{preds/city_vis_13} 
\input{preds/city_vis_2} 
\input{preds/city_vis_3} 
\input{preds/city_vis_28} 
\input{preds/city_vis_4} 
\input{preds/city_vis_11} 
\input{preds/palette}
\vspace{-0.15cm}
\caption{Example predictions showing better panoptic segmentation for thing (\emph{car}) and stuff (\emph{wall}, \emph{sign}, \emph{light}) classes on 
SYNTHIA $\rightarrow$ Cityscapes.}
\label{fig:pred_car}
\end{figure*}

\begin{figure*}[h!]
\centering
\input{preds/prediction_head}
\input{preds/city_vis_1} 
\input{preds/city_vis_22} 
\input{preds/city_vis_30} 
\input{preds/city_vis_29} 
\input{preds/city_vis_21} 
\input{preds/palette}
\vspace{-0.15cm}
\caption{Example predictions showing better panoptic segmentation for \emph{bus}, \emph{traffic sign} and \emph{traffic light} on SYNTHIA $\rightarrow$ Cityscapes.}
\label{fig:pred_miscellaneous}
\end{figure*}

\begin{figure*}[h!]
\centering
\input{preds/prediction_head}
\input{preds/map_vis_3}
\input{preds/map_vis_5}
\input{preds/map_vis_1}
\input{preds/map_vis_4}
\input{preds/map_vis_2}
\input{preds/palette}
\vspace{-0.15cm}
\caption{Example predictions on SYNTHIA $\rightarrow$ Mapillary Vistas.}
\label{fig:pred_set8}
\end{figure*}

\begin{figure*}[h!]
\centering
\input{preds/prediction_head2}
\input{preds/map_vis_9} 
\input{preds/map_vis_10} 
\input{preds/map_vis_11} 
\input{preds/map_vis_8}
\input{preds/map_vis_6}
\input{preds/map_vis_7}
\input{preds/palette}
\vspace{-0.15cm}
\caption{Example predictions on SYNTHIA $\rightarrow$ Mapillary Vistas.}
\label{fig:pred_set9}
\end{figure*}

\begin{figure*}[h!]
\centering
\input{preds/prediction_head2}
\input{preds/map_vis_16} 
\input{preds/map_vis_12} 
\input{preds/map_vis_17} 
\input{preds/map_vis_13} 
\input{preds/map_vis_14} 
\input{preds/map_vis_15} 
\input{preds/palette}
\vspace{-0.15cm}
\caption{Example predictions on SYNTHIA $\rightarrow$ Mapillary Vistas.}
\label{fig:pred_set10}
\end{figure*}

\begin{figure*}[h!]
\centering
\input{preds/c2fcs} 
\input{preds/palette}
\vspace{-0.15cm}
\caption{Visual prediction results of EDAPS on Cityscapes $\rightarrow$ Foggy Cityscapes.}
\label{fig:c2fcs}
\end{figure*}

\begin{figure*}[h!]
\centering
\input{preds/edaps_vs_mdecbu_01_comp_new} 
\input{preds/palette}
\vspace{-0.15cm}
\caption{Visual comparison of EDAPS and M-Dec-BU (baseline) predictions on SYNTHIA $\rightarrow$ Cityscapes.}
\label{fig:edaps_vs_mdecbu_01}
\end{figure*}

%% file: tables/edaps_sota_c2fc.tex
\begin{tabular}{l @{\quad} cccccccccccccccc @{\quad} c @{\quad} c @{\quad} c}
\toprule 
UDA Method & \rots{road} & \rots{sidewalk\quads\quads} & \rots{building} & \rots{wall} & \rots{fence} & \rots{pole} & \rots{light} & \rots{sign} & \rots{veg} & \rots{sky} & \rots{person} & \rots{rider} & \rots{car} & \rots{bus} & \rots{m.bike} & \rots{bike} & mSQ & mRQ & mPQ \\
\midrule
UniDAPS-Baseline~\cite{carion2020detr} &92.5&48.9&60.6&6.0&10.7&5.3&9.9&23.6&49.7&55.6&22.3&15.4&38.5&23.7&1.6&2.8&70.0&38.6&29.2 \\
 DAF~\cite{chen2018wild}&\B{94.0}&{54.5}&57.7&6.7&10.0&7.0&6.6&25.5&44.6&{59.1}&26.7&16.7&42.2&36.6&4.5&16.9&70.6&41.7&31.8 \\
 FDA~\cite{yang2020fda}&93.8& 53.1& 62.2& 8.2& 13.4& 7.3& 7.6& {28.9}& 50.8& 49.7&25.0& 22.6& 42.9& 36.3& 10.3& 15.2& 71.4&43.5& 33.0\\
 AdvEnt~\cite{vu2019advent}&93.8&52.7&56.3&5.7&13.5&{10.0}&{10.9}&27.7&40.7&57.9&27.8&29.4&44.7&28.6&11.6&20.8&72.3&43.7&33.3 \\
 CRST~\cite{zou2019confidence}&91.8&49.7&66.1&6.4&14.5&5.2&8.6&21.5&56.3&50.7&30.5&30.7&46.3&34.2&11.7&22.1&72.2&44.9&34.1 \\
 SVMin~\cite{guan2021scale}&93.4& 53.4& 62.2& {12.3}& 15.5& 7.0& 8.5& 18.0& 54.3& 57.1& 31.2& 29.6& 45.2& 35.6& 11.5& 22.7&72.4&45.5& 34.8\\
  \midrule
 CVRN~\cite{huang2021cross}&93.6&52.3&{65.3}&7.5&{15.9}&5.2&7.4&22.3&{57.8}&48.7&32.9&30.9&49.6&38.9&18.0&25.2&72.7&46.7&35.7 \\
UniDAPS \cite{zhang2022hierarchical}&{93.9}& 53.1& 63.9& 8.7& 14.0& 3.8&10.0& 26.0& 53.5& 49.6 &{38.0}& {35.4}& {57.5}& {44.2}& {28.9}& {29.8}&{72.9}&{49.5}&{37.6} \\
\midrule
\B{EDAPS w/ MiT-B2 (Ours)}& 90.3 & 64.8 & 80.0 & 20.7 & \B{32.0} & 47.9 & 45.4 & 63.3 & 85.1 & \B{71.8} & 46.8 & 48.0 & 64.0 & \B{52.6} & 34.1 & 36.2 & 78.9 & 68.7 & 55.1 \\
\B{EDAPS w/ MiT-B5 (Ours)} &  91.0 & \B{68.5} & \B{80.9} & \B{24.1} 
& 29.0 & \B{50.1} & \B{47.2} & \B{67.0} & \B{85.3} & \B{71.8} & \B{50.9} & \B{51.2} & \B{64.7} & 47.7 & \B{36.9} & \B{41.5} & \B{79.2} & \B{70.5} & \B{56.7} \\
\bottomrule
\end{tabular}

%% file: tables/edaps_sota_c2m.tex
\begin{tabular}{l @{\quad} cccccccccccccccc @{\quad} c @{\quad} c @{\quad} c}
\toprule 
UDA Method & \rots{road} & \rots{sidewalk\quads\quads} & \rots{building} & \rots{wall} & \rots{fence} & \rots{pole} & \rots{light} & \rots{sign} & \rots{veg} & \rots{sky} & \rots{person} & \rots{rider} & \rots{car} & \rots{bus} & \rots{m.bike} & \rots{bike} & mSQ & mRQ & mPQ \\
\midrule
CRST~\cite{zou2019confidence} &77.0 &22.6 &40.2 &7.8 &10.5 &5.5 &11.3 &21.8 &56.5 &77.6 &29.4 &18.4 &56.0 &27.7 &11.9 &18.4 &72.4 &39.9 &30.8 \\
FDA~\cite{yang2020fda} &74.3 &{23.4} &42.3 &9.6 &11.2 &6.4 &{15.4} &23.5 &60.4 &78.5 &33.9 &19.9 &52.9 &8.4 &{17.5} &16.0 &72.3 &40.3 &30.9 \\
AdvEnt~\cite{vu2019advent}  &76.2 &20.5 &42.6 &6.8 &9.4 &4.6 &12.7 &24.1 &59.9 &83.1 &34.1 &{22.9} &54.1 &16.0 &13.5 &18.6 &72.7 &40.3 &31.2 \\
{CVRN~\cite{huang2021cross}} &\B{77.3} &21.0 &{47.8} &{10.5} &{13.4} &{7.5} &14.1 &{25.1} &{62.1} &\B{86.4} &\B{37.7} &20.4 &{55.0} &{21.7} &14.3 &{21.4} &{73.8} &{42.8} &{33.5} \\
\midrule
\B{EDAPS w/ MiT-B5 (Ours)} &  58.8 & \B{43.4} & \B{57.1} & \B{25.6} & \B{29.1} & \B{34.3} & \B{35.5} & \B{41.2} & \B{77.8} & 59.1 & 35.0 & \B{23.8} & \B{56.7} & \B{36.0} & \B{24.3} & \B{25.5} & \B{75.9} & \B{53.4} & \B{41.2} \\
\bottomrule
\end{tabular}

%% file: tables/adversarial_train_analysis.tex
\begin{tabular}{lccccc}
\toprule
UDA Method                      & mAP           & mIoU  & mSQ   & mRQ       & mPQ \\
\midrule
EDAPS Adversarial Train                 & 23.6	         & 39.4	       & 63.8	       & 35.0          &	26.2 \\
EDAPS Self Train (Ours)        & \textbf{34.4} & \textbf{57.5} & \textbf{72.7} & \textbf{53.6} & \textbf{41.2} \\
\bottomrule
\end{tabular}

%% file: tables/mask-rcnn-ablation.tex
\begin{tabular}{cccccc @{\quad} ccccc}
\toprule 
&
$\mathcal{L}_\mathrm{RPN\text{-}cls}$
& $\mathcal{L}_\mathrm{RPN\text{-}box}$
& $\mathcal{L}_\mathrm{RoI\text{-}cls}$
& $\mathcal{L}_\mathrm{RoI\text{-}box}$
& $\mathcal{L}_\mathrm{RoI\text{-}mask}$
& mAP 
& mIoU 
& mSQ 
& mRQ 
& mPQ \\
\midrule
Model 1
&
& 
& \checkmark
& \checkmark
& \checkmark
& 9.3\spm{4.4} 
& 57.5\spm{0.4} 
& 70.9\spm{7.3} 
& 43.2\spm{5.8} 
& 30.8\spm{1.3} 
\\
Model 2
&
\checkmark
& 
& \checkmark
& \checkmark
& \checkmark
& 4.8\spm{2.1} 
& 57.9\spm{1.1} 
& 62.9\spm{2.4} 
& 38.8\spm{0.3} 
& 29.6\spm{0.2} 
\\
Model 3
&
& \checkmark
& \checkmark
& \checkmark
& \checkmark
& 16.9\spm{5.9} 
& 57.8\spm{0.4} 
& 72.5\spm{1.6} 
& 45.1\spm{3.2} 
& 34.7\spm{2.4} 
\\
Model 4
&
\checkmark
& \checkmark
& 
& 
& \checkmark
& 0.5\spm{0.3} 
& 57.3\spm{0.4} 
& 45.5\spm{0.1} 
& 38.4\spm{0.7} 
& 29.5\spm{0.5} 
\\
Model 5
&
\checkmark
& \checkmark
& 
& \checkmark
& \checkmark
& 2.3\spm{2.0} 
& 57.9\spm{0.5} 
& 45.6\spm{0.1} 
& 38.4\spm{0.4} 
& 29.5\spm{0.3}
\\
Model 6
&
\checkmark
& \checkmark
& \checkmark
& 
& \checkmark
& 29.6\spm{0.4} 
& 57.5\spm{0.4} 
& 71.7\spm{0.5} 
& 50.4\spm{0.7} 
& 38.2\spm{0.7} 
\\
Model 7
&
\checkmark
& \checkmark
& \checkmark
& \checkmark
& 
& 9.7\spm{1.9} 
& 57.0\spm{1.3} 
& 65.3\spm{3.8} 
& 43.7\spm{1.6} 
& 32.7\spm{0.9} 
\\
Model 8
&
\checkmark
& \checkmark
& \checkmark
& \checkmark
& \checkmark
& 34.4\spm{0.5} 
& 57.5\spm{0.0} 
& 72.7\spm{0.2} 
& 53.6\spm{0.5} 
& 41.2\spm{0.4} 
\\

\bottomrule
\end{tabular}

%% file: preds/prediction_head.tex
{\footnotesize
\begin{tabularx}{\linewidth}{*{4}{Y}}
Image & CVRN~\cite{huang2021cross} & EDAPS (Ours) & Ground Truth \\
\end{tabularx}
} %

%% file: preds/palette.tex
\scriptsize
\setlength\tabcolsep{1pt}
{
\newcolumntype{P}[1]{>{\centering\arraybackslash}p{#1}}
\begin{tabular}{@{}*{20}{P{0.09\columnwidth}}@{}}
     {\cellcolor[rgb]{0.5,0.25,0.5}}\textcolor{white}{road} 
     &{\cellcolor[rgb]{0.957,0.137,0.91}}sidew. 
     &{\cellcolor[rgb]{0.275,0.275,0.275}}\textcolor{white}{build.} 
     &{\cellcolor[rgb]{0.4,0.4,0.612}}\textcolor{white}{wall} 
     &{\cellcolor[rgb]{0.745,0.6,0.6}}fence 
     &{\cellcolor[rgb]{0.6,0.6,0.6}}pole 
     &{\cellcolor[rgb]{0.98,0.667,0.118}}tr. light
     &{\cellcolor[rgb]{0.863,0.863,0}}tr. sign 
     &{\cellcolor[rgb]{0.42,0.557,0.137}}veget. 
     &{\cellcolor[rgb]{0.596,0.984,0.596}}terrain 
     &{\cellcolor[rgb]{0.275,0.510,0.706}}sky
     &{\cellcolor[rgb]{0.863,0.078,0.235}}\textcolor{white}{person} 
     &{\cellcolor[rgb]{1,0,0}}\textcolor{white}{rider} 
     &{\cellcolor[rgb]{0,0,0.557}}\textcolor{white}{car} 
     &{\cellcolor[rgb]{0,0,0.275}}\textcolor{white}{truck} 
     &{\cellcolor[rgb]{0,0.235,0.392}}\textcolor{white}{bus}
     &{\cellcolor[rgb]{0,0.392,0.471}}\textcolor{white}{train} 
     &{\cellcolor[rgb]{0,0,0.902}}\textcolor{white}{m.bike} 
     & {\cellcolor[rgb]{0.467,0.043,0.125}}\textcolor{white}{bike}
     &{\cellcolor[rgb]{0,0,0}}\textcolor{white}{n/a.}
\end{tabular}
}

%% file: preds/prediction_head2.tex
{\footnotesize
\begin{tabularx}{\linewidth}{*{4}{Y}}
Image & Source-Only & EDAPS (Ours) & Ground Truth \\
\end{tabularx}
} %

%% file: main.bbl
\begin{thebibliography}{10}\itemsep=-1pt

\bibitem{araslanov2021self}
Nikita Araslanov and Stefan Roth.
\newblock Self-supervised augmentation consistency for adapting semantic segmentation.
\newblock In {\em Proceedings of the IEEE/CVF Conference on Computer Vision and Pattern Recognition}, pages 15384--15394, 2021.

\bibitem{carion2020end}
Nicolas Carion, Francisco Massa, Gabriel Synnaeve, Nicolas Usunier, Alexander Kirillov, and Sergey Zagoruyko.
\newblock End-to-end object detection with transformers.
\newblock In {\em European Conference on Computer Vision}, pages 213--229. Springer, 2020.

\bibitem{carion2020detr}
Nicolas Carion, Francisco Massa, Gabriel Synnaeve, Nicolas Usunier, Alexander Kirillov, and Sergey Zagoruyko.
\newblock End-to-end object detection with transformers.
\newblock In {\em European Conference on Computer Vision}, pages 213--229. Springer, 2020.

\bibitem{chen2017deeplab}
Liang-Chieh Chen, George Papandreou, Iasonas Kokkinos, Kevin Murphy, and Alan~L Yuille.
\newblock Deeplab: Semantic image segmentation with deep convolutional nets, atrous convolution, and fully connected crfs.
\newblock {\em IEEE Transactions on Pattern Analysis and Machine Intelligence}, 40(4):834--848, 2017.

\bibitem{chen2018domain}
Yuhua Chen, Wen Li, Christos Sakaridis, Dengxin Dai, and Luc Van~Gool.
\newblock Domain adaptive faster r-cnn for object detection in the wild.
\newblock In {\em Proceedings of the IEEE/CVF Conference on Computer Vision and Pattern Recognition}, pages 3339--3348, 2018.

\bibitem{chen2018wild}
Yuhua Chen, Wen Li, Christos Sakaridis, Dengxin Dai, and Luc Van~Gool.
\newblock Domain adaptive faster r-cnn for object detection in the wild.
\newblock In {\em Proceedings of the IEEE conference on computer vision and pattern recognition}, pages 3339--3348, 2018.

\bibitem{chen2018road}
Yuhua Chen, Wen Li, and Luc Van~Gool.
\newblock Road: Reality oriented adaptation for semantic segmentation of urban scenes.
\newblock In {\em Proceedings of the IEEE Conference on Computer Vision and Pattern Recognition}, pages 7892--7901, 2018.

\bibitem{chen2021scale}
Yuhua Chen, Haoran Wang, Wen Li, Christos Sakaridis, Dengxin Dai, and Luc Van~Gool.
\newblock Scale-aware domain adaptive faster r-cnn.
\newblock {\em International Journal of Computer Vision}, 129(7):2223--2243, 2021.

\bibitem{cheng2019panoptic}
Bowen Cheng, Maxwell~D Collins, Yukun Zhu, Ting Liu, Thomas~S Huang, Hartwig Adam, and Liang-Chieh Chen.
\newblock Panoptic-deeplab: A simple, strong, and fast baseline for bottom-up panoptic segmentation.
\newblock In {\em Proceedings of the IEEE/CVF Conference on Computer Vision and Pattern Recognition}, pages 12475--12485, 2020.

\bibitem{cheng2021per}
Bowen Cheng, Alex Schwing, and Alexander Kirillov.
\newblock Per-pixel classification is not all you need for semantic segmentation.
\newblock {\em Advances in Neural Information Processing Systems}, 34:17864--17875, 2021.

\bibitem{choi2019self}
Jaehoon Choi, Taekyung Kim, and Changick Kim.
\newblock Self-ensembling with gan-based data augmentation for domain adaptation in semantic segmentation.
\newblock In {\em Proceedings of the IEEE/CVF International Conference on Computer Vision}, pages 6830--6840, 2019.

\bibitem{chollet2017xception}
Fran{\c{c}}ois Chollet.
\newblock Xception: Deep learning with depthwise separable convolutions.
\newblock In {\em Proceedings of the IEEE/CVF Conference on Computer Vision and Pattern Recognition}, pages 1251--1258, 2017.

\bibitem{cordts2016cityscapes}
Marius Cordts, Mohamed Omran, Sebastian Ramos, Timo Rehfeld, Markus Enzweiler, Rodrigo Benenson, Uwe Franke, Stefan Roth, and Bernt Schiele.
\newblock The cityscapes dataset for semantic urban scene understanding.
\newblock In {\em Proceedings of the IEEE/CVF Conference on Computer Vision and Pattern Recognition}, pages 3213--3223, 2016.
\newblock Dataset URL: \url{https://www.cityscapes-dataset.com/}, Dataset License: \url{https://www.cityscapes-dataset.com/license/}.

\bibitem{dosovitskiy2020image}
Alexey Dosovitskiy, Lucas Beyer, Alexander Kolesnikov, Dirk Weissenborn, Xiaohua Zhai, Thomas Unterthiner, Mostafa Dehghani, Matthias Minderer, Georg Heigold, Sylvain Gelly, et~al.
\newblock An image is worth 16x16 words: Transformers for image recognition at scale.
\newblock {\em International Conference on Learning Representations}, 2021.

\bibitem{du2019ssf}
Liang Du, Jingang Tan, Hongye Yang, Jianfeng Feng, Xiangyang Xue, Qibao Zheng, Xiaoqing Ye, and Xiaolin Zhang.
\newblock Ssf-dan: Separated semantic feature based domain adaptation network for semantic segmentation.
\newblock In {\em Proceedings of the IEEE/CVF International Conference on Computer Vision}, pages 982--991, 2019.

\bibitem{ganin2016domain}
Yaroslav Ganin, Evgeniya Ustinova, Hana Ajakan, Pascal Germain, Hugo Larochelle, Fran{\c{c}}ois Laviolette, Mario Marchand, and Victor Lempitsky.
\newblock Domain-adversarial training of neural networks.
\newblock {\em The Journal of Machine Learning Research}, 17(1):2096--2030, 2016.

\bibitem{george2018encoder}
Liang-Chieh Chen Yukun~Zhu George, Papandreou~Florian Schroff, and Hartwig Adam.
\newblock Encoder-decoder with atrous separable convolution for semantic image segmentation.
\newblock {\em Proceedings of the European Conference on Computer Vision}, 2018.

\bibitem{gong2021dlow}
Rui Gong, Wen Li, Yuhua Chen, Dengxin Dai, and Luc Van~Gool.
\newblock Dlow: Domain flow and applications.
\newblock {\em International Journal of Computer Vision}, 129(10):2865--2888, 2021.

\bibitem{guan2021scale}
Dayan Guan, Jiaxing Huang, Shijian Lu, and Aoran Xiao.
\newblock Scale variance minimization for unsupervised domain adaptation in image segmentation.
\newblock {\em Pattern Recognition}, 112:107764, 2021.

\bibitem{he2017mask}
Kaiming He, Georgia Gkioxari, Piotr Doll{\'a}r, and Ross Girshick.
\newblock Mask r-cnn.
\newblock In {\em Proceedings of the IEEE/CVF International Conference on Computer Vision}, pages 2961--2969, 2017.

\bibitem{hoffman2018cycada}
Judy Hoffman, Eric Tzeng, Taesung Park, Jun-Yan Zhu, Phillip Isola, Kate Saenko, Alexei Efros, and Trevor Darrell.
\newblock Cycada: Cycle-consistent adversarial domain adaptation.
\newblock In {\em International Conference on Machine Learning}, pages 1989--1998. PMLR, 2018.

\bibitem{hoffman2016fcns}
Judy Hoffman, Dequan Wang, Fisher Yu, and Trevor Darrell.
\newblock Fcns in the wild: Pixel-level adversarial and constraint-based adaptation.
\newblock {\em arXiv preprint arXiv:1612.02649}, 2016.

\bibitem{hoyer2021daformer}
Lukas Hoyer, Dengxin Dai, and Luc Van~Gool.
\newblock {DAFormer}: Improving network architectures and training strategies for domain-adaptive semantic segmentation.
\newblock In {\em Proceedings of the IEEE/CVF Conference on Computer Vision and Pattern Recognition}, 2022.

\bibitem{hoyer2022hrda}
Lukas Hoyer, Dengxin Dai, and Luc Van~Gool.
\newblock {HRDA}: Context-aware high-resolution domain-adaptive semantic segmentation.
\newblock In {\em Proceedings of the European Conference on Computer Vision (ECCV)}, 2022.

\bibitem{hoyer2023domain}
Lukas Hoyer, Dengxin Dai, and Luc Van~Gool.
\newblock Domain adaptive and generalizable network architectures and training strategies for semantic image segmentation.
\newblock {\em arXiv preprint arXiv:2304.13615}, 2023.

\bibitem{hoyer2023mic}
Lukas Hoyer, Dengxin Dai, Haoran Wang, and Luc Van~Gool.
\newblock {MIC}: Masked image consistency for context-enhanced domain adaptation.
\newblock In {\em Proceedings of the IEEE/CVF Conference on Computer Vision and Pattern Recognition}, 2023.

\bibitem{hoyer2021improving}
Lukas Hoyer, Dengxin Dai, Qin Wang, Yuhua Chen, and Luc Van~Gool.
\newblock Improving semi-supervised and domain-adaptive semantic segmentation with self-supervised depth estimation.
\newblock {\em International Journal of Computer Vision}, 2021.

\bibitem{huang2021cross}
Jiaxing Huang, Dayan Guan, Aoran Xiao, and Shijian Lu.
\newblock Cross-view regularization for domain adaptive panoptic segmentation.
\newblock In {\em Proceedings of the IEEE/CVF Conference on Computer Vision and Pattern Recognition}, pages 10133--10144, 2021.

\bibitem{kamann2021benchmarking}
Christoph Kamann and Carsten Rother.
\newblock Benchmarking the robustness of semantic segmentation models with respect to common corruptions.
\newblock {\em International Journal of Computer Vision}, 129(2):462--483, 2021.

\bibitem{kang2020pixel}
Guoliang Kang, Yunchao Wei, Yi Yang, Yueting Zhuang, and Alexander Hauptmann.
\newblock Pixel-level cycle association: A new perspective for domain adaptive semantic segmentation.
\newblock {\em Advances in Neural Information Processing Systems}, 33:3569--3580, 2020.

\bibitem{kim2020learning}
Myeongjin Kim and Hyeran Byun.
\newblock Learning texture invariant representation for domain adaptation of semantic segmentation.
\newblock In {\em Proceedings of the IEEE/CVF Conference on Computer Vision and Pattern Recognition}, pages 12975--12984, 2020.

\bibitem{kirillov2019panopticFP}
Alexander Kirillov, Ross Girshick, Kaiming He, and Piotr Doll{\'a}r.
\newblock Panoptic feature pyramid networks.
\newblock In {\em Proceedings of the IEEE/CVF Conference on Computer Vision and Pattern Recognition}, pages 6399--6408, 2019.

\bibitem{kirillov2019panoptic}
Alexander Kirillov, Kaiming He, Ross Girshick, Carsten Rother, and Piotr Doll{\'a}r.
\newblock Panoptic segmentation.
\newblock In {\em Proceedings of the IEEE/CVF Conference on Computer Vision and Pattern Recognition}, pages 9404--9413, 2019.

\bibitem{lee2013pseudo}
Dong-Hyun Lee.
\newblock Pseudo-label: The simple and efficient semi-supervised learning method for deep neural networks.
\newblock In {\em Int. Conf. Mach. Learn. Worksh.}, 2013.

\bibitem{li2019bidirectional}
Yunsheng Li, Lu Yuan, and Nuno Vasconcelos.
\newblock Bidirectional learning for domain adaptation of semantic segmentation.
\newblock In {\em Proceedings of the IEEE/CVF Conference on Computer Vision and Pattern Recognition}, pages 6936--6945, 2019.

\bibitem{li2022cross}
Yu-Jhe Li, Xiaoliang Dai, Chih-Yao Ma, Yen-Cheng Liu, Kan Chen, Bichen Wu, Zijian He, Kris Kitani, and Peter Vajda.
\newblock Cross-domain adaptive teacher for object detection.
\newblock In {\em Proceedings of the IEEE/CVF Conference on Computer Vision and Pattern Recognition}, pages 7581--7590, 2022.

\bibitem{li2022panoptic}
Zhiqi Li, Wenhai Wang, Enze Xie, Zhiding Yu, Anima Anandkumar, Jose~M Alvarez, Ping Luo, and Tong Lu.
\newblock Panoptic segformer: Delving deeper into panoptic segmentation with transformers.
\newblock In {\em Proceedings of the IEEE/CVF Conference on Computer Vision and Pattern Recognition}, pages 1280--1289, 2022.

\bibitem{pdam}
Dongnan Liu, Donghao Zhang, Yang Song, Fan Zhang, Lauren O’Donnell, Heng Huang, Mei Chen, and Weidong Cai.
\newblock Pdam: A panoptic-level feature alignment framework for unsupervised domain adaptive instance segmentation in microscopy images.
\newblock {\em IEEE Transactions on Medical Imaging}, 40(1):154--165, 2021.

\bibitem{liu2021bapa}
Yahao Liu, Jinhong Deng, Xinchen Gao, Wen Li, and Lixin Duan.
\newblock Bapa-net: Boundary adaptation and prototype alignment for cross-domain semantic segmentation.
\newblock In {\em Proceedings of the IEEE/CVF International Conference on Computer Vision}, pages 8801--8811, 2021.

\bibitem{liu2021learning}
Yunan Liu, Shanshan Zhang, Yang Li, and Jian Yang.
\newblock Learning to adapt via latent domains for adaptive semantic segmentation.
\newblock {\em Advances in Neural Information Processing Systems}, 34, 2021.

\bibitem{long2015learning}
Mingsheng Long, Yue Cao, Jianmin Wang, and Michael Jordan.
\newblock Learning transferable features with deep adaptation networks.
\newblock In {\em International Conference on Machine Learning}, pages 97--105. PMLR, 2015.

\bibitem{long2018conditional}
Mingsheng Long, Zhangjie Cao, Jianmin Wang, and Michael~I Jordan.
\newblock Conditional adversarial domain adaptation.
\newblock {\em Neural Information Processing Systems}, 31, 2018.

\bibitem{loshchilov2018decoupled}
Ilya Loshchilov and Frank Hutter.
\newblock Decoupled weight decay regularization.
\newblock In {\em International Conference on Learning Representations}, 2018.

\bibitem{luo2019taking}
Yawei Luo, Liang Zheng, Tao Guan, Junqing Yu, and Yi Yang.
\newblock Taking a closer look at domain shift: Category-level adversaries for semantics consistent domain adaptation.
\newblock In {\em Proceedings of the IEEE/CVF Conference on Computer Vision and Pattern Recognition}, pages 2507--2516, 2019.

\bibitem{mei2020instance}
Ke Mei, Chuang Zhu, Jiaqi Zou, and Shanghang Zhang.
\newblock Instance adaptive self-training for unsupervised domain adaptation.
\newblock In {\em Proceedings of the European Conference on Computer Vision}, pages 415--430. Springer, 2020.

\bibitem{melas2021pixmatch}
Luke Melas-Kyriazi and Arjun~K Manrai.
\newblock Pixmatch: Unsupervised domain adaptation via pixelwise consistency training.
\newblock In {\em Proceedings of the IEEE/CVF Conference on Computer Vision and Pattern Recognition}, pages 12435--12445, 2021.

\bibitem{mohan2021efficientps}
Rohit Mohan and Abhinav Valada.
\newblock Efficientps: Efficient panoptic segmentation.
\newblock {\em International Journal of Computer Vision}, 129(5):1551--1579, 2021.

\bibitem{naseer2021intriguing}
Muzammal Naseer, Kanchana Ranasinghe, Salman Khan, Munawar Hayat, Fahad~Shahbaz Khan, and Ming-Hsuan Yang.
\newblock Intriguing {Properties} of {Vision} {Transformers}.
\newblock In {\em Neural Information Processing Systems}, 2021.

\bibitem{neuhold2017mapillary}
Gerhard Neuhold, Tobias Ollmann, Samuel Rota~Bulo, and Peter Kontschieder.
\newblock The mapillary vistas dataset for semantic understanding of street scenes.
\newblock In {\em Proceedings of the IEEE international conference on computer vision}, pages 4990--4999, 2017.

\bibitem{olsson2021classmix}
Viktor Olsson, Wilhelm Tranheden, Juliano Pinto, and Lennart Svensson.
\newblock Classmix: Segmentation-based data augmentation for semi-supervised learning.
\newblock In {\em Proceedings of the IEEE/CVF Winter Conference on Applications of Computer Vision}, pages 1369--1378, 2021.

\bibitem{pan2019transferrable}
Yingwei Pan, Ting Yao, Yehao Li, Yu Wang, Chong-Wah Ngo, and Tao Mei.
\newblock Transferrable prototypical networks for unsupervised domain adaptation.
\newblock In {\em Proceedings of the IEEE/CVF Conference on Computer Vision and Pattern Recognition}, pages 2239--2247, 2019.

\bibitem{paszke2017automatic}
Adam Paszke, Sam Gross, Soumith Chintala, Gregory Chanan, Edward Yang, Zachary DeVito, Zeming Lin, Alban Desmaison, Luca Antiga, and Adam Lerer.
\newblock Automatic differentiation in pytorch.
\newblock In {\em NIPS Autodiff Workshop}, 2017.

\bibitem{porzi2019seamless}
Lorenzo Porzi, Samuel~Rota Bulo, Aleksander Colovic, and Peter Kontschieder.
\newblock Seamless scene segmentation.
\newblock In {\em Proceedings of the IEEE/CVF Conference on Computer Vision and Pattern Recognition}, pages 8277--8286, 2019.

\bibitem{richter2016playing}
Stephan~R Richter, Vibhav Vineet, Stefan Roth, and Vladlen Koltun.
\newblock Playing for data: Ground truth from computer games.
\newblock In {\em European Conference on Computer Vision}, pages 102--118. Springer, 2016.

\bibitem{ros2016synthia}
German Ros, Laura Sellart, Joanna Materzynska, David Vazquez, and Antonio~M Lopez.
\newblock The synthia dataset: A large collection of synthetic images for semantic segmentation of urban scenes.
\newblock In {\em Proceedings of the IEEE/CVF Conference on Computer Vision and Pattern Recognition}, pages 3234--3243, 2016.
\newblock Dataset URL: \url{http://synthia-dataset.net/}, Dataset License: CC BY-NC-SA 3.0.

\bibitem{saha2021learning}
Suman Saha, Anton Obukhov, Danda~Pani Paudel, Menelaos Kanakis, Yuhua Chen, Stamatios Georgoulis, and Luc Van~Gool.
\newblock Learning to relate depth and semantics for unsupervised domain adaptation.
\newblock In {\em Proceedings of the IEEE/CVF Conference on Computer Vision and Pattern Recognition}, pages 8197--8207, 2021.

\bibitem{saito2019strong}
Kuniaki Saito, Yoshitaka Ushiku, Tatsuya Harada, and Kate Saenko.
\newblock Strong-weak distribution alignment for adaptive object detection.
\newblock In {\em Proceedings of the IEEE/CVF Conference on Computer Vision and Pattern Recognition}, pages 6956--6965, 2019.

\bibitem{saito2018maximum}
Kuniaki Saito, Kohei Watanabe, Yoshitaka Ushiku, and Tatsuya Harada.
\newblock Maximum classifier discrepancy for unsupervised domain adaptation.
\newblock In {\em Proceedings of the IEEE/CVF Conference on Computer Vision and Pattern Recognition}, pages 3723--3732, 2018.

\bibitem{sajjadi2016regularization}
Mehdi Sajjadi, Mehran Javanmardi, and Tolga Tasdizen.
\newblock Regularization with stochastic transformations and perturbations for deep semi-supervised learning.
\newblock {\em Advances in Neural Information Processing Systems}, 29, 2016.

\bibitem{sakaridis2018model}
Christos Sakaridis, Dengxin Dai, Simon Hecker, and Luc Van~Gool.
\newblock Model adaptation with synthetic and real data for semantic dense foggy scene understanding.
\newblock In {\em Proceedings of the European Conference on Computer Vision}, pages 687--704, 2018.

\bibitem{sohn2020fixmatch}
Kihyuk Sohn, David Berthelot, Nicholas Carlini, Zizhao Zhang, Han Zhang, Colin~A Raffel, Ekin~Dogus Cubuk, Alexey Kurakin, and Chun-Liang Li.
\newblock Fixmatch: Simplifying semi-supervised learning with consistency and confidence.
\newblock {\em Advances in Neural Information Processing Systems}, 33:596--608, 2020.

\bibitem{sun2022safe}
Tao Sun, Cheng Lu, Tianshuo Zhang, and Haibin Ling.
\newblock Safe self-refinement for transformer-based domain adaptation.
\newblock In {\em Proceedings of the IEEE/CVF Conference on Computer Vision and Pattern Recognition}, pages 7191--7200, 2022.

\bibitem{tarvainen2017mean}
Antti Tarvainen and Harri Valpola.
\newblock Mean teachers are better role models: Weight-averaged consistency targets improve semi-supervised deep learning results.
\newblock {\em Advances in Neural Information Processing Systems}, 30, 2017.

\bibitem{tranheden2021dacs}
Wilhelm Tranheden, Viktor Olsson, Juliano Pinto, and Lennart Svensson.
\newblock Dacs: Domain adaptation via cross-domain mixed sampling.
\newblock In {\em Proceedings of the IEEE/CVF Winter Conference on Applications of Computer Vision}, pages 1379--1389, 2021.

\bibitem{tsai2018learning}
Yi-Hsuan Tsai, Wei-Chih Hung, Samuel Schulter, Kihyuk Sohn, Ming-Hsuan Yang, and Manmohan Chandraker.
\newblock Learning to adapt structured output space for semantic segmentation.
\newblock In {\em Proceedings of the IEEE/CVF Conference on Computer Vision and Pattern Recognition}, pages 7472--7481, 2018.

\bibitem{tsai2019domain}
Yi-Hsuan Tsai, Kihyuk Sohn, Samuel Schulter, and Manmohan Chandraker.
\newblock Domain adaptation for structured output via discriminative patch representations.
\newblock In {\em Proceedings of the IEEE/CVF International Conference on Computer Vision}, pages 1456--1465, 2019.

\bibitem{vaswani2017attention}
Ashish Vaswani, Noam Shazeer, Niki Parmar, Jakob Uszkoreit, Llion Jones, Aidan~N Gomez, {\L}ukasz Kaiser, and Illia Polosukhin.
\newblock Attention is all you need.
\newblock {\em Advances in Neural Information Processing Systems}, 30, 2017.

\bibitem{vu2019advent}
Tuan-Hung Vu, Himalaya Jain, Maxime Bucher, Matthieu Cord, and Patrick P{\'e}rez.
\newblock Advent: Adversarial entropy minimization for domain adaptation in semantic segmentation.
\newblock In {\em Proceedings of the IEEE/CVF Conference on Computer Vision and Pattern Recognition}, pages 2517--2526, 2019.

\bibitem{wang2020classes}
Haoran Wang, Tong Shen, Wei Zhang, Ling-Yu Duan, and Tao Mei.
\newblock Classes matter: A fine-grained adversarial approach to cross-domain semantic segmentation.
\newblock In {\em Proceedings of the European Conference on Computer Vision}, pages 642--659. Springer, 2020.

\bibitem{wang2021domain}
Qin Wang, Dengxin Dai, Lukas Hoyer, Luc Van~Gool, and Olga Fink.
\newblock Domain adaptive semantic segmentation with self-supervised depth estimation.
\newblock In {\em Proceedings of the IEEE/CVF International Conference on Computer Vision}, pages 8515--8525, 2021.

\bibitem{wang2021pyramid}
Wenhai Wang, Enze Xie, Xiang Li, Deng-Ping Fan, Kaitao Song, Ding Liang, Tong Lu, Ping Luo, and Ling Shao.
\newblock Pyramid vision transformer: A versatile backbone for dense prediction without convolutions.
\newblock In {\em Proceedings of the IEEE/CVF International Conference on Computer Vision}, pages 568--578, 2021.

\bibitem{wang2020differential}
Zhonghao Wang, Mo Yu, Yunchao Wei, Rogerio Feris, Jinjun Xiong, Wen-mei Hwu, Thomas~S Huang, and Honghui Shi.
\newblock Differential treatment for stuff and things: A simple unsupervised domain adaptation method for semantic segmentation.
\newblock In {\em Proceedings of the IEEE/CVF Conference on Computer Vision and Pattern Recognition}, pages 12635--12644, 2020.

\bibitem{xie2021segformer}
Enze Xie, Wenhai Wang, Zhiding Yu, Anima Anandkumar, Jose~M Alvarez, and Ping Luo.
\newblock Segformer: Simple and efficient design for semantic segmentation with transformers.
\newblock {\em Advances in Neural Information Processing Systems}, 34:12077--12090, 2021.

\bibitem{xiong2019upsnet}
Yuwen Xiong, Renjie Liao, Hengshuang Zhao, Rui Hu, Min Bai, Ersin Yumer, and Raquel Urtasun.
\newblock Upsnet: A unified panoptic segmentation network.
\newblock In {\em Proceedings of the IEEE/CVF Conference on Computer Vision and Pattern Recognition}, pages 8818--8826, 2019.

\bibitem{xu2020cross}
Minghao Xu, Hang Wang, Bingbing Ni, Qi Tian, and Wenjun Zhang.
\newblock Cross-domain detection via graph-induced prototype alignment.
\newblock In {\em Proceedings of the IEEE/CVF Conference on Computer Vision and Pattern Recognition}, pages 12355--12364, 2020.

\bibitem{xu2021cdtrans}
Tongkun Xu, Weihua Chen, Pichao Wang, Fan Wang, Hao Li, and Rong Jin.
\newblock Cdtrans: Cross-domain transformer for unsupervised domain adaptation.
\newblock {\em arXiv preprint arXiv:2109.06165}, 2021.

\bibitem{yang2020fda}
Yanchao Yang and Stefano Soatto.
\newblock Fda: Fourier domain adaptation for semantic segmentation.
\newblock In {\em Proceedings of the IEEE/CVF Conference on Computer Vision and Pattern Recognition}, pages 4085--4095, 2020.

\bibitem{yu2020gradient}
Tianhe Yu, Saurabh Kumar, Abhishek Gupta, Sergey Levine, Karol Hausman, and Chelsea Finn.
\newblock Gradient surgery for multi-task learning.
\newblock {\em Advances in Neural Information Processing Systems}, 33:5824--5836, 2020.

\bibitem{zhang2022hierarchical}
Jingyi Zhang, Jiaxing Huang, and Shijian Lu.
\newblock Hierarchical mask calibration for unified domain adaptive panoptic segmentation.
\newblock {\em arXiv preprint arXiv:2206.15083}, 2022.

\bibitem{zhang2021prototypical}
Pan Zhang, Bo Zhang, Ting Zhang, Dong Chen, Yong Wang, and Fang Wen.
\newblock Prototypical pseudo label denoising and target structure learning for domain adaptive semantic segmentation.
\newblock In {\em Proceedings of the IEEE/CVF Conference on Computer Vision and Pattern Recognition}, pages 12414--12424, 2021.

\bibitem{zhang2019category}
Qiming Zhang, Jing Zhang, Wei Liu, and Dacheng Tao.
\newblock Category anchor-guided unsupervised domain adaptation for semantic segmentation.
\newblock {\em Advances in neural information processing systems}, 32, 2019.

\bibitem{zheng2021rethinking}
Sixiao Zheng, Jiachen Lu, Hengshuang Zhao, Xiatian Zhu, Zekun Luo, Yabiao Wang, Yanwei Fu, Jianfeng Feng, Tao Xiang, Philip~HS Torr, et~al.
\newblock Rethinking semantic segmentation from a sequence-to-sequence perspective with transformers.
\newblock In {\em Proceedings of the IEEE/CVF Conference on Computer Vision and Pattern Recognition}, pages 6881--6890, 2021.

\bibitem{zhou2021context}
Qianyu Zhou, Zhengyang Feng, Qiqi Gu, Jiangmiao Pang, Guangliang Cheng, Xuequan Lu, Jianping Shi, and Lizhuang Ma.
\newblock Context-aware mixup for domain adaptive semantic segmentation.
\newblock {\em IEEE Transactions on Circuits and Systems for Video Technology}, 2022.

\bibitem{zou2018unsupervised}
Yang Zou, Zhiding Yu, BVK Kumar, and Jinsong Wang.
\newblock Unsupervised domain adaptation for semantic segmentation via class-balanced self-training.
\newblock In {\em Proceedings of the European Conference on Computer Vision}, pages 289--305, 2018.

\bibitem{zou2019confidence}
Yang Zou, Zhiding Yu, Xiaofeng Liu, BVK Kumar, and Jinsong Wang.
\newblock Confidence regularized self-training.
\newblock In {\em Proceedings of the IEEE/CVF International Conference on Computer Vision}, pages 5982--5991, 2019.

\end{thebibliography}
